\documentclass[twoside,11pt]{article}

\usepackage{jmlr2e}

\usepackage{amsmath,amssymb,xcolor,algorithmic,algorithm}
\usepackage[latin1]{inputenc}
\usepackage[english]{babel}
\usepackage{dsfont}

\newtheorem{assumption}[theorem]{Assumption}
\usepackage[colorlinks=false,allbordercolors={1 1 1}]{hyperref}

\usepackage{macros}

\newcommand{\Kinf}{\cK_{\text{inf}}}
\newcommand{\Tgoesto}{\underset{T\rightarrow \infty}{\rightarrow}}

\jmlrheading{16}{2016}{1-42}{7/14; Revised 2/15}{1/16}{E. Kaufmann, O. Capp\'e and A. Garivier}

\ShortHeadings{Complexity of Best-Arm Identification in Multi-Armed Bandits}{Kaufmann, Capp\'e and Garivier}
\firstpageno{1}

\begin{document}

\title{On the Complexity of Best-Arm Identification in Multi-Armed Bandit Models}

\author{\name Emilie Kaufmann \email emilie.kaufmann@telecom-paristech.fr \\
       \addr LTCI, CNRS, T\'el\'ecom ParisTech\\
       \addr 46, rue Barrault, 75013 Paris
       \AND
       \name Olivier Capp\'e \email olivier.cappe@telecom-paristech.fr \\
       \addr LTCI, CNRS, T\'el\'ecom ParisTech\\
       \addr 46, rue Barrault, 75013 Paris
      \AND
       \name Aur\'elien Garivier \email aurelien.garivier@math.univ-toulouse.fr \\
       \addr Institut de Math\'ematiques de Toulouse ; UMR5219 \\
	\addr Universit\'e de Toulouse ; CNRS \\
	\addr UPS IMT, F-31062 Toulouse Cedex 9
      }

\editor{G\'abor Lugosi}

\maketitle

\begin{abstract}
  The stochastic multi-armed bandit model is a simple abstraction that has proven useful in many different contexts in statistics and machine learning. Whereas the achievable limit in terms of regret minimization is now well known, our aim is to contribute to a better understanding of the performance in terms of identifying the $m$ best arms. We introduce generic notions of complexity for the two dominant frameworks considered in the literature: fixed-budget and fixed-confidence settings. In the fixed-confidence setting, we provide the first known distribution-dependent lower bound on the complexity that involves information-theoretic quantities and holds when $m\geq1$ under general assumptions. In the specific case of two armed-bandits, we derive refined lower bounds in both the fixed-confidence and fixed-budget settings, along with matching algorithms for Gaussian and Bernoulli bandit models. These results show in particular that the complexity of the fixed-budget setting may be smaller than the complexity 
of the fixed-confidence setting, contradicting the familiar behavior observed when testing fully specified alternatives. In addition, we also provide improved sequential stopping rules that have guaranteed error probabilities and shorter average running times. The proofs rely on two technical results that are of independent interest : a deviation lemma for self-normalized sums (Lemma~\ref{thm:subgaussian}) and a novel change of measure inequality for bandit models (Lemma~\ref{lem:Cornerstone}).
\end{abstract}


\maketitle

\section{Introduction}

We investigate in this paper the complexity of finding the $m$ best arms in a
stochastic multi-armed bandit model. A bandit model $\nu$ is a collection of
$K$ arms, where each arm $\nu_a \; (1\leq a\leq K)$ is a probability
distribution on $\R$ with expectation $\mu_a$.  At each time $t=1,2,\dots$, an
agent chooses an option $A_t\in\{1,\dots,K\}$ and receives an independent draw
$Z_t$ from the corresponding arm $\nu_{A_t}$. We denote by $\bP_\nu$
(resp. $\bE_\nu$) the probability law (resp. expectation) of the process
$(Z_t)$. The agent's goal is to identify the $m$ best arms, that is, the set
$\cS^*_m$ of indices of the $m$ arms with highest expectation. Letting
$(\mu_{[1]},\dots, \mu_{[K]})$ be the K-tuple of expectations
$(\mu_1,\dots,\mu_K)$ sorted in decreasing order, we assume that the bandit
model $\nu$ belongs to a class $\cM_m$ such that for every $\nu\in\cM_m$,
$\mu_{[m]}>\mu_{[m+1]}$, so that $\cS^*_m$ is unambiguously defined.

In order to identify $\cS^*_m$, the agent must use a strategy defining which
arms to sample from, when to stop sampling, and which set $\hat{S}_m$ to
choose. More precisely, its strategy consists in a triple $\cA=((A_t),\tau,\hat{S}_m)$ in which :
\begin{itemize}
\item the \textit{sampling rule} determines, based on past observations,
which arm $A_t$ is chosen at time $t$; in other words, $A_t$ is
$\cF_{t-1}$-measurable, with $\mathcal{F}_t=\sigma(A_1,Z_1,\dots,A_t,Z_t)$;
\item the \textit{stopping rule} $\tau$ controls the end of the data acquisition phase
and is a stopping time with respect to $(\mathcal{F}_{t})_{t\in\N}$ satisfying
$\bP(\tau < + \infty)=1$; 
\item the \textit{recommendation rule} provides the
arm selection and is a $\mathcal{F}_\tau$-measurable random subset $\hat{S}_m$ of
$\{1,\dots,K\}$ of size $m$.
\end{itemize}

In the bandit literature, two different settings have been considered.  In the
\emph{fixed-confidence setting}, a risk parameter $\delta$ is fixed, and a
strategy $\cA(\delta)$ is called \emph{$\delta$-PAC} if, for every choice of
$\nu\in\mathcal{M}_m$, $\bP_\nu(\hat{S}_m = \cS^*_m)\geq 1-\delta$. {The goal is to obtain $\delta$-PAC strategies 
that require a  number of draws $\tau_\delta$ that is as small as possible. More precisely, we focus on strategies minimizing the expected number 
of draws $\bE_\nu[\tau_\delta]$, which is also called the \emph{sample complexity}. The subscript $\delta$ in $\tau_\delta$ will be omitted when there is no ambiguity. We call a family of strategies $A=(\cA(\delta))_{\delta \in (0,1)}$ $PAC$ if for every $\delta$, $\cA(\delta)$ is $\delta$-PAC.}

Alternatively, in the \emph{fixed-budget setting}, the number of draws $\tau$ is fixed in advance 
to some value $t\in \N$ and for this budget $t$, the goal is to choose the sampling and
recommendation rules of a strategy $\cA(t)$ so as to minimize the failure
probability $p_t(\nu) := \bP_\nu(\hat{S}_m \neq S^*_m)$.  In the fixed-budget
setting, a family of strategies A=$(\cA(t))_{t\in\N^*}$ is called \emph{consistent} if, for every choice of
$\nu\in\mathcal{M}_m$, $p_t(\nu)$ tends to zero when $t$ increases to infinity.

In order to unify and compare these approaches, we define the \emph{complexity}
$\kappa_C(\nu)$ (resp. $\kappa_B(\nu)$) of best-arm identification in the
fixed-confidence (resp. fixed-budget) setting as follows:
{
\begin{equation}
  \kappa_{\text{C}}(\nu) = \inf_{\emph{A} \ \text{PAC}} \limsup_{\delta
    \rightarrow 0} \frac{\bE_\nu[\tau_{\delta}]}{\log \frac{1}{\delta}}, \hspace{1cm}
  \kappa_{\text{B}}(\nu)= \inf_{\emph{A} \ \text{consistent}} \left(\limsup_{t
      \rightarrow \infty} -\frac{1}{t}\log p_t(\nu)\right)^{-1}. \label{def:Complexities}
\end{equation}}

Heuristically, on the one hand for a given bandit model $\nu$, and a small value of
$\delta$, a fixed-confidence optimal strategy needs an average number of
samples of order $\kappa_{C}(\nu)\log \frac{1}{\delta}$ to identify the $m$
best arms with probability at least $1-\delta$. On the other hand,
 for large values of $t$ the probability of error of a fixed-budget optimal strategy is 
of order $\exp(-\kappa_B(\nu) t )$, which means that a 
budget of approximately $t=\kappa_{B}(\nu)\log \frac{1}{\delta}$ draws is required to ensure a probability
of error of order $\delta$. 
Most of the existing performance bounds for the
fixed confidence and fixed budget settings can indeed be expressed using these
complexity measures.

In this paper, we aim at evaluating and comparing these two complexities.  To
achieve this, two ingredients are needed: a lower bound on the sample
complexity of any $\delta$-PAC algorithm (resp. on the failure probability of
any consistent algorithm) and a $\delta$-PAC (resp. consistent) strategy whose
sample complexity (resp. failure probability) attains the lower bound (often
referred to as a 'matching' strategy).  We present below new lower bounds on
$\kappa_C(\nu)$ and $\kappa_B(\nu)$ that feature information-theoretic
quantities as well as strategies that match these lower bounds in
two-armed bandit models.

A particular class of algorithms will be considered in the following: those using
a \textit{uniform sampling strategy}, that sample the arms in a round-robin
fashion.  Whereas it is well known that when $K>2$ uniform sampling is not
desirable, it will prove efficient in some examples of two-armed bandits. This
specific setting, relevant in practical applications discussed in
Section~\ref{sec:2arms}, is studied in greater details along the paper. In this
case, an algorithm using uniform sampling can be regarded as a statistical test
of the hypothesis $H_0 : (\mu_1 \leq \mu_2)$ against $H_1 : (\mu_1>\mu_2)$
based on paired samples ($X_s,Y_s$) of $\nu_1,\nu_2$; namely a test based on a
fixed number of samples in the fixed-budget setting, and, a \textit{sequential
  test} in the fixed-confidence setting, in which a (random) stopping rule
determines when the experiment is to be terminated.

Classical sequential testing theory provides a first element of comparison
between the fixed-budget and fixed-confidence settings, in the simpler case of
fully specified alternatives.  Consider for instance the case where $\nu_1$ and
$\nu_2$ are Gaussian laws with the same known variance $\sigma^2$, the means
$\mu_1$ and $\mu_2$ being known up to a permutation. Denoting by $P$ the joint
distribution of the paired samples $(X_s,Y_s)$, one must choose between the
hypotheses $H_0:P=\norm{\mu_1}{\sigma^2}\otimes \norm{\mu_2}{\sigma^2}$ and
$H_1:P=\norm{\mu_2}{\sigma^2}\otimes \norm{\mu_1}{\sigma^2}$. It is known since
\cite{Wald45SPRT} that among the sequential tests such that type I and type II
error probabilities are both smaller than $\delta$, the Sequential Probability
Ratio Test (SPRT) minimizes the expected number of required samples, and is
such that $\bE_\nu[\tau]\simeq2\sigma^2/(\mu_1-\mu_2)^2\log(1/\delta)$. However, the
batch test that minimizes both probabilities of error is the Likelihood Ratio
test; it can be shown to require a sample size of order
$8\sigma^2/(\mu_1-\mu_2)^2\log(1/\delta)$ in order to ensure that both type I
and type II error probabilities are smaller than $\delta$. Thus, when the
sampling strategy is uniform and the parameters are known, there is a clear
gain in using randomized stopping strategies.  We will show below that this
conclusion is not valid anymore when the values of $\mu_1$ and $\mu_2$ are not
assumed to be known. Indeed, for two-armed Gaussian bandit models we show that
$\kappa_B(\nu)=\kappa_C(\nu)$ and for two-armed Bernoulli bandit models we show
that $\kappa_C(\nu)>\kappa_B(\nu)$.

\subsection{Related Works}

Bandit models have received a considerable interest since their introduction
by~\cite{Thompson33} in the context of medical trials.  An important focus was
set on a different perspective, in which each observation is considered as a
reward: the agent aims at maximizing its cumulative rewards.  Equivalently, his
goal is to minimize the expected \emph{regret} up to horizon $t\geq 1$ defined as $R_t(\nu) =
t\mu_{[1]} - \bE_\nu\left[\sum_{s=1}^{t} Z_s\right]\;.$ Regret minimization,
which is paradigmatic of the so-called \emph{exploration versus exploitation
  dilemma}, was introduced by \cite{Robbins:Freq52} and its complexity is well
understood for simple families of parametric bandits.  In generic one-parameter
models, \cite{LaiRobbins85bandits} prove that, with a proper notion of
consistency adapted to regret minimization,
\[\inf_{A \ \text{consistent}}\liminf_{t\rightarrow \infty}\frac{R_t(\nu)}{\log
  t} \geq \sum_{a : \mu_{a}<\mu_{[1]}} \frac{(\mu_{[1]} -
  \mu_a)}{\text{KL}(\nu_a,\nu_{[1]})}\;,\] where $\text{KL}(\nu_i,\nu_j)$
denotes the Kullback-Leibler divergence between distributions $\nu_i$ and
$\nu_j$.  This bound was later generalized by \cite{Burn:Kat96} to
distributions that depend on several parameters.  Since then, non-asymptotic
analyses of efficient algorithms matching this bound have been proposed.
Optimal algorithms include the KL-UCB algorithm of \cite{KLUCB:Journal}---a
variant of UCB1 \citep{Auer:al02} using informational upper bounds, Thompson
Sampling \citep{ALT12,AG:AISTAT13}, the DMED algorithm \citep{HondaTakemura11}
and Bayes-UCB~\citep{AISTATS12}.  This paper is a contribution towards
similarly characterizing the complexity of \emph{pure exploration}, where the
goal is to determine the best arms without trying to maximize the cumulative
observations.

\cite{Bubeck:al11} show that in the fixed-budget setting, when $m=1$, any sampling
strategy designed to minimize regret performs poorly with respect to the
\textit{simple regret} $r_t := \mu^* - \mu_{\hat{S}_1}$, which is closely
related to the probability $p_t(\nu)$ of recommending the wrong arm. Therefore,
good strategies for best-arm identification need to be quite different from
regret-minimizing strategies. We will show below that the complexities $\kappa_B(\nu)$ and $\kappa_C(\nu)$ of
best-arm identification also involve information terms, but these are different from the Kullback-Leibler divergence featured in Lai and Robbins' lower bound on the regret.

The problem of best-arm identification has been studied since the 1950s under the name
`ranking and identification problems'.  The first advances on this topic are
summarized in the monograph by \cite{Bechofer:al68} who consider the
fixed-confidence setting and strategies based on uniform sampling. In the fixed
confidence setting, \cite{Paulson:64} first introduces a sampling strategy
based on eliminations for single best arm identification: the arms are
successively discarded, the remaining arms being sampled uniformly. This idea
was later used for example by \cite{Jennison:al84,MaronMoore:97} and by
\cite{EvenDar:al06} in the context of bounded bandit models, in which each arm
$\nu_a$ is a probability distribution on $[0,1]$. $m$ best arms identification
with $m>1$ was considered for example by \cite{HeidrichMeisner:al09}, in the
context of reinforcement learning. \cite{Shivaram:al12} later proposed {an algorithm that is 
no longer based on eliminations, called LUCB (for Lower and Upper Confidence Bounds) and still
designed for bounded bandit models}. Bounded
distributions are in fact particular examples of distributions with subgaussian
tails, to which the proposed algorithms can be easily generalized.  A relevant
quantity introduced in the analysis of algorithms for bounded (or subgaussian)
bandit models is the `complexity term'
\begin{equation}
  H(\nu)= \sum_{a \in \{1, 2, \dots
    K\}}\frac{1}{\Delta_a^2}
  \ \ \ \  \text{with}  \ \ \Delta_{a} = 
  \begin{cases}
    \mu_a - \mu_{[m+1]} & \text{for} \ a\in \mathcal{S}^*_m,\\
    \mu_{[m]} - \mu_{a} & \text{for} \ a\in (\mathcal{S}^*_m)^c.
  \end{cases} \label{def:HComplexity}
\end{equation}

The upper bound on the sample complexity of the LUCB algorithm of
\cite{Shivaram:al12} implies in particular that $\kappa_{C}(\nu) \leq
292H(\nu)$. Some of the existing works on the fixed-confidence setting do not
bound $\tau$ in expectation but rather show that $\bP_\nu(\hat{\cS}_m=\cS_m^*,
\tau = O\left(H(\nu)\right))\geq 1- \delta$.  These results are not directly
comparable with the complexity $\kappa_C(\nu)$, although no significant gap is
to be observed yet.

Recent works have focused on obtaining upper bounds on the number of samples whose dependency in terms of the squared-gaps $\Delta_a$ (for subgaussian arms) is optimal when the $\Delta_a$'s go to zero, and $\delta$ remains fixed. \cite{Karnin:al13} and \cite{Jamiesonal14LILUCB} exhibit $\delta$-PAC algorithms for which there exists a constant $C$ such that, with high probability, the number of samples used satisfies  
\[\tau \leq C_0 \sum_{a\neq a^*}\frac{1}{\Delta_a^2}\log\left(\frac{1}{\delta}\log\frac{1}{\Delta_a}\right),\]
and \cite{Jamiesonal14LILUCB} show that the dependency in $\Delta_a^{-2}\log(\log(\Delta_a^{-1}))$ is optimal when $\Delta_a$ goes to zero. However, the constant $C_0$ is large 
and does not lead to improved upper bounds on the complexity term $\kappa_C(\nu)$.

For $m=1$, the work of \cite{MannorTsi:04} provides a lower bound on
$\kappa_C(\nu)$ in the case of Bernoulli bandit models, under the following
$\epsilon$-relaxation sometimes considered in the literature. For some
tolerance parameter $\epsilon\geq 0$ the agent has to ensure that $\hat{S}_m$
is included in the set of $(\epsilon,m)$ optimal arms $\cS^*_{m,\epsilon} = \{a
: \mu_a \geq \mu_{[m]}-\epsilon \}$ with probability at least $1-\delta$. {This relaxation has to be considered, for example, when $\mu_{[m]}=\mu_{[m+1]}$, but has never been considered in the literature for the fixed-budget setting. In
this paper, we focus on the case $\epsilon=0$ that allows for a comparison between the fixed-confidence and fixed-budget settings.} 
\cite{MannorTsi:04} show that if an algorithm is
$\delta$-PAC, then in the bandit model $\nu=(\cB(\mu_1),\dots,\cB(\mu_K))$ such
that $\forall a$, $\mu_a \in [0,\alpha]$ for some $\alpha \in (0,1)$, there
exists {two sets $\cG_\alpha(\nu)\subset\cS_1^*$ and $\cH_\alpha(\nu)\subset
\{1,\dots,K\}\backslash\cS_1^*$} and a positive constant $C_\alpha$ such that
\[\bE_\nu[\tau] \geq C_\alpha \left(\sum_{a\in
    \cG_\alpha(\nu)}\frac{1}{\epsilon^2} + \sum_{a \in
    \cH_\alpha(\nu)}\frac{1}{(\mu_{[1]} -
    \mu_a)^2}\right)\log\left(\frac{1}{8\delta}\right).\] This bound is non
asymptotic (as emphasized by the authors), although not completely explicit. In
particular, the subset $\cG_\alpha$ and $\cH_\alpha$ do not always form a
partition of the arms (it can happen that $\cG_\alpha \cup
\cH_\alpha \neq \{1,\dots,K\}$), hence the complexity term does not involve a sum over all the
arms.  For $m>1$, the only lower bound available in the literature is the
worst-case result of~\cite{Shivaram:al12}. It states that for every
$\delta$-PAC algorithm \textit{there exists} a bandit model $\nu$ such that
$\bE_\nu[\tau] \geq {K}/{(18375\epsilon^2)}\log\left({m}/{8\delta}\right)$.
This result, however, does not provide a lower bound on the complexity $\kappa_C(\nu)$.

The fixed-budget setting has been studied by
\cite{Bubeck:BestArm10,Bubeck:al11} for single best-arm identification in
bounded bandit models. For multiple arm identification ($m>1$), still in
bounded bandit models, \cite{Bubeck:alMult13} introduce the SAR (for Successive
Accepts and Rejects) algorithm. An upper bound on the failure probability of
the SAR algorithm yields $\kappa_{B}(\nu) \leq 8\log(K)H\left(\nu\right)$.

For $m=1$, \cite{Bubeck:BestArm10} prove an asymptotic lower bound on the
probability of error
for Bernoulli bandit models. They state that for every algorithm and every
bandit problem $\nu$ such that $\forall a$, $\mu_1\in[\alpha,1-\alpha]$, there
exists a permutation of the arms $\nu'$ such that
\[ p_t(\nu')\geq \exp(-{t}/{C_\alpha H_2(\nu'))}), \ \  \text{with} \ \ \
H_2(\nu) = \max_{i: \mu_{[i]}<\mu_{[1]}}\frac{i}{\left(\mu_{[1]}-\mu_{[i]}\right)^{2}} \ \ \text{and} \ \ C_\alpha = \frac{\alpha(1-\alpha)}{5 + o(1)}.
\]

\cite{Gabillon:al12} propose the UGapE algorithm for $m$ best-arm
identification for $m>1$. By changing only one parameter in some confidence
regions, this algorithm can be adapted either to the fixed-budget or to the
fixed-confidence setting. However, a careful inspection
shows that UGapE cannot be used in the fixed-budget setting without the
knowledge of the complexity term $H(\nu)$. This drawback is shared by other
algorithms designed for the fixed-budget setting, like the UCB-E algorithm of
\cite{Bubeck:BestArm10} or the KL-LUCB-E algorithm of \cite{COLT13}.

\subsection{Content of the Paper}

The gap between lower and upper bounds known so far does not permit to identify
exactly the complexity terms $\kappa_B(\nu)$ and $\kappa_C(\nu)$ defined in
(\ref{def:Complexities}). Not only do they involve imprecise multiplicative
constants but by analogy with the Lai and Robbins' bound for the expected regret, the quantities $H(\nu),H_2(\nu)$ presented above are only expected to be relevant in the Gaussian case.

The improvements of this paper mainly concern the fixed-confidence setting,
which will be considered in the next three Sections. We first propose in
Section 2 a distribution-dependent lower bound on $\kappa_C(\nu)$ that holds
for $m>1$ and for general classes of bandit models (Theorem
\ref{thm:GeneralBoundFC}). This information-theoretic lower bound permits to
interpret the quantity $H(\nu)$ defined in (\ref{def:HComplexity}) as a
subgaussian approximation.

Theorem \ref{thm:2armsFC} in Section \ref{sec:2arms} proposes a tighter lower
bound on $\kappa_C(\nu)$ for general classes of two-armed bandit models, as
well as a lower bound on the sample complexity of $\delta$-PAC algorithms using
uniform sampling. In Section \ref{sec:Examples} we propose, for Gaussian
bandits with known---but possibly different---variances, an algorithm exactly
matching this bound. We also consider the case of Bernoulli distributed arms,
for which we show that uniform sampling is nearly optimal in most cases.  We
propose a new algorithm using uniform sampling and a non-trivial stopping
strategy that is close to matching the lower bound.

Section \ref{sec:FixedBudget} gathers our contributions to the fixed-budget
setting. For two-armed bandits, Theorem \ref{thm:2armsFB} provides a lower
bound on $\kappa_B(\nu)$ that is in general different from the lower bound
obtained for $\kappa_C(\nu)$ in the fixed-confidence setting. Then we propose
matching algorithms for the fixed-budget setting that allow for a comparison
between the two settings. For Gaussian bandits, we show that
$\kappa_C(\nu)=\kappa_B(\nu)$, whereas for Bernoulli bandits
$\kappa_C(\nu)>\kappa_B(\nu)$, proving that the two complexities are not
necessarily equal. As a first step towards a lower bound on $\kappa_B(\nu)$
when $m>1$, we also give in Section \ref{sec:FixedBudget} new lower bounds on
the probability of error $p_t(\nu)$ of any consistent algorithm, for Gaussian
bandit models.

Section \ref{sec:Experiments} contains numerical experiments that illustrate the
performance of matching algorithms for Gaussian and Bernoulli two-armed bandits, comparing the fixed-confidence and fixed-budget settings.

Our contributions follow from two main mathematical results of more general interest. Lemma
\ref{lem:Cornerstone} provides a general relation between the expected number
of draws and Kullback-Leibler divergences of the arms' distributions, which is
the key element to derive the lower bounds (it also permits, for example, to derive Lai and Robbin's lower bound on the regret in a few lines). 
Lemma \ref{thm:subgaussian} is a tight deviation inequality for martingales with sub-Gaussian increments, in the spirit of the Law of Iterated Logarithm, that permits here to derive efficient
matching algorithms for two-armed bandits.
 
 \section{Generic Lower Bound in the Fixed-Confidence
  Setting\label{sec:GeneralBoundFC}}

Introducing the Kullback-Leibler divergence of any two probability
distributions $p$ and $q$:
$$\K(p,q)  = 
\left\{
  \begin{array}{l}
    \int \log \left[\frac{dp}{dq}(x)\right]dp(x) \ \text{if} \ q \ll p,\\
    + \infty \ \text{otherwise},
  \end{array}
\right.
$$
we make the assumption that there exists a set $\cP$ of probability measures
such that for all $\nu=(\nu_1,\dots,\nu_K)\in \cM_m$, for $a\in\{1,\dots,K\},
\nu_a\in\cP$ and that $\cP$ satisfies
\[
\forall p,q\in\cP, \ p\neq q \ \Rightarrow \ 0< \K(p,q) < + \infty.
\]
A class $\cM_m$ of bandit models satisfying this property is called
\textit{identifiable}.

All the distribution-dependent lower bounds derived in the bandit literature
\citep[e.g.,][]{LaiRobbins85bandits,MannorTsi:04,Bubeck:BestArm10} rely on
\textit{changes of distribution}, and so do ours. A change of distribution
relates the probabilities of the same event under two different bandit models
$\nu$ and $\nu'$. The following lemma 
provides a new,
synthetic, inequality from which lower bounds are directly derived. This result, proved in Appendix \ref{proofs:ChangeDist}, encapsulates the technical aspects of the change of distribution. 
The main ingredient in its proof is a lower bound on the \emph{expected log-likelihood ratio} of the observations under two different bandit models which 
is of interest on its own and is stated as Lemma~\ref{lem:CornerstoneLike} in Appendix~\ref{proofs:ChangeDist}.
To illustrate the interest of Lemma~\ref{lem:Cornerstone} even beyond the pure exploration framework, we give in Appendix
\ref{proof:LaiRobbins} a new, simple proof of \cite{Burn:Kat96}'s generalization
of Lai and Robbins' lower bound in the regret minimization framework based on Lemma \ref{lem:Cornerstone}.

Let $N_a(t) = \sum_{s=1}^t \ind_{\{A_s=a\}}$ be the number of draws of arm $a$ between the instants 1 and $t$ and
$N_a=N_a(\tau)$ be the total number of draws of arm $a$ by some algorithm
$\cA=((A_t),\tau,\hat{S}_m)$.

\begin{lemma} \label{lem:Cornerstone} Let $\nu$ and $\nu'$ be two bandit
  models with $K$ arms such that for all $a$, the distributions $\nu_a$ and $\nu_a'$ are mutually absolutely continuous. 
  For any almost-surely finite stopping time $\sigma$ with respect to $(\cF_t)$,
  \[
  \sum_{a=1}^{K} \bE_\nu[N_a(\sigma)] \K(\nu_a,\nu'_a)\geq
  \sup_{\cE \in \cF_\sigma} \ d(\bP_\nu(\cE),\bP_{\nu'}(\cE)),
  \]
  where $d(x,y):= x\log (x/y) + (1-x)\log((1-x)/(1-y))$ is the binary relative entropy, with the convention 
  that $d(0,0)=d(1,1)=0$.
\end{lemma}

\begin{remark}
This result can be considered as a generalization of Pinsker's inequality to bandit models: in combination with the inequality $d(p,q)\geq 2(p-q)^2$, it yields:
  \[
  \sup_{\cE \in \cF_\sigma} \left|\bP_\nu(\cE)-\bP_{\nu'}(\cE)\right| \leq 
  \sqrt{\frac{\sum_{a=1}^{K} \bE_\nu[N_a(\sigma)] \K(\nu_a,\nu'_a)}{2}}\;.
  \]

However, it is important in this paper \emph{not to use this weaker form} of the statement, as we will consider events $\cE$ of probability very close to $0$ or $1$. In this regime, we will make use of the following inequality:
\begin{equation}\label{TrickKL}
\forall \ x \in [0,1], \ \ \ \ d(x,1-x) \geq \log\frac{1}{2.4 x}\;,
\end{equation}
which can be checked easily. 
\end{remark}

\subsection{Lower Bound on the Sample Complexity of a $\delta$-PAC Algorithm}

We now propose a non-asymptotic lower bound on the expected number of samples
needed to identify the $m$ best arms in the fixed confidence setting, which
straightforwardly yields a lower bound on $\kappa_C(\nu)$.

Theorem \ref{thm:GeneralBoundFC} holds for an identifiable class of bandit
models of the form:
\begin{equation}\cM_m = \{\nu=(\nu_{1},\dots,\nu_{K}) : \nu_i \in \cP ,
  \mu_{[m]}>\mu_{[m+1]}\}\label{BM:General}\end{equation}
such that the set of probability measures $\cP$ satisfies Assumption~\ref{assumption} below. 

\begin{assumption}\label{assumption} For all $p,q \in \cP^2$ such that $p\neq q$,
for all $\alpha>0$,

there exists $q_1\in\cP$: $\K(p,q) < \K(p,q_1)<\K(p,q)+\alpha$
and $\bE_{X\sim q_1}[X]>\bE_{X\sim q}[X]$,

there exists $q_2\in\cP$: $\K(p,q) < \K(p,q_2)<\K(p,q)+\alpha$
and $\bE_{X\sim q_2}[X]<\bE_{X\sim q}[X]$.
\end{assumption} 

\medskip

These continuity conditions are reminiscent of the assumptions of
\cite{LaiRobbins85bandits}; they include families of parametric bandits
continuously parameterized by their means (e.g., Bernoulli, Poisson, exponential distributions).

\begin{theorem}\label{thm:GeneralBoundFC} Let $\nu\in \cM_m$, where $\cM_m$ is defined by (\ref{BM:General}), 
  and assume that $\cP$ satisfies Assumption~\ref{assumption}; any algorithm that is $\delta$-PAC on
  $\cM_m$ satisfies, for $\delta \leq 0.15$,
  \[\bE_\nu[\tau]\geq \left[\sum_{a\in
      \mathcal{S}^*_m}\frac{1}{\K(\nu_{a},\nu_{[m+1]})} + \sum_{a\notin
      \mathcal{S}^*_m}\frac{1}{\K(\nu_{a},\nu_{[m]})}\right]\log\left(\frac{1}{2.4\delta}\right).\]
\end{theorem}

\textbf{Proof.} Without loss of generality, one may assume that the arms are
ordered such that $\mu_{1}\geq \dots \geq \mu_{K}$.  Thus
$\mathcal{S}^*_m=\{1,...,m\}$. Let $\cA=((A_t),\tau,\hat{S}_m)$ be a
$\delta$-PAC algorithm and fix $\alpha>0$. For all $a\in\{1,\dots,K\}$, from
Assumption~\ref{assumption} there exists an alternative model
\[\nu' = (\nu_1,\dots,\nu_{a-1},\nu_{a}',\nu_{a+1},\dots,\nu_K)\]
in which the only arm modified is arm $a$, and $\nu_a'$ is such that:
\begin{itemize}
\item $\K(\nu_a,\nu_{m+1})<\K(\nu_a,\nu_a')<\K(\nu_a,\nu_{m+1})+\alpha$ and
  $\mu_a'<\mu_{m+1}$ if $a\in \{1,\dots,m\}$,
\item $\K(\nu_a,\nu_{m})<\K(\nu_a,\nu_a')<\K(\nu_a,\nu_{m})+\alpha$ and
  $\mu_a'>\mu_{m}$ if $a\in \{m+1,\dots,K\}$.
\end{itemize}
In particular, on the bandit model $\nu'$ the set of optimal arms is no longer
$\{1,\dots,m\}$. Thus, introducing the event $\cE=(\hat{\cS}_m=\{1,\dots,m\})\in
\cF_\tau$, any $\delta$-PAC algorithm satisfies $\bP_\nu(\cE)\geq 1-\delta$ and
$\bP_{\nu'}(\cE)\leq \delta$.  Lemma \ref{lem:Cornerstone} applied to the
stopping time $\tau$ (such that $N_a(\tau)=N_a$ is the total number of draws of
arm $a$) and the monotonicity properties of $d(x,y)$ ($x\mapsto d(x,y)$ is increasing when $x>y$ and decreasing when $x<y$)
yield
\[\K(\nu_{a},\nu'_a)\bE_\nu[N_a] \geq d(1-\delta,\delta)\geq \log(1/2.4\delta),\]
where the last inequality follows from \eqref{TrickKL}. 
From the definition of the alternative model, one obtains for
$a\in\{1,\dots,m\}$ or $b\in\{m+1,\dots,K\}$ respectively, for every
$\alpha>0$,
\[\bE_\nu[N_a] \geq \frac{\log(1/2.4\delta)}{\K(\nu_a,\nu_{m+1})+\alpha}
\ \ \ \ \text{and} \ \ \ \ \bE_\nu[N_b] \geq
\frac{\log(1/2.4\delta)}{\K(\nu_b,\nu_{m})+\alpha}.\]  
Letting $\alpha$ tend to zero and summing over the arms yields the bound on $\bE_\nu[\tau]=\sum_{a=1}^{K}\bE_\nu[N_a]$.

\begin{remark} 
This inequality can be made tighter for values of $\delta$ that are sufficiently close to zero, for which the right-hand-side can then be made arbitrarily close to $\log(1/\delta)$.

  Lemma \ref{lem:Cornerstone} can also be used to improve the result
  of \cite{MannorTsi:04} that holds for $m=1$ under the $\epsilon$-relaxation
  described before.  Combining the changes of distribution of this paper with
  Lemma \ref{lem:Cornerstone} yields, for every $\epsilon>0$ and
  $\delta\leq0.15$,
  \[\bE_\nu[\tau] \geq \left(\frac{|\{ a : \mu_a \geq \mu_{[1]} -
      \epsilon \}| -1 }{\K\left(\cB({\mu_{[1]}}) , \cB({\mu_{[1]} -
          \epsilon})\right)} + \sum_{a : \mu_a \leq \mu_{[1]} - \epsilon}
    \frac{1}{\K\left(\cB({\mu_a}),\cB({\mu_{[1]} +\epsilon})\right)}\right)
  \log \frac{1}{2.4\delta},\] where $ |\cX|$ denotes the cardinal of the set
  $\cX$ and $\cB(\mu)$ the Bernoulli distribution of mean $\mu$.
\end{remark}

\subsection{Bounds on the Complexity for Exponential Bandit Models}

Theorem \ref{thm:GeneralBoundFC} yields the following lower bound on the
complexity term:
\[\kappa_C(\nu) \geq \sum_{a\in\cS^*_m}\frac{1}{\K(\nu_a,\nu_{[m+1]})} +
\sum_{a\notin\cS^*_m}\frac{1}{\K(\nu_a,\nu_{[m]})}.\] Thus, one may want to
obtain strategies whose sample complexity can be proved to be of the same
magnitude.  The only algorithm that has been analyzed so far with an
information-theoretic perspective is the KL-LUCB algorithm of \cite{COLT13},
designed for \textit{exponential bandit models}: that is
\[\cM_m=\left\{\nu=(\nu_{\theta_1},\dots,\nu_{\theta_K}) :
  (\theta_1,\dots,\theta_K)\in \Theta^K, \theta_{[m]}>\theta_{[m+1]}\right\},\]
where $\nu_{\theta}$ belongs to a \textit{canonical one-parameter exponential
  family}. {This means that there exists a twice differentiable strictly convex function $b$ such that
$\nu_\theta$ has a density with respect to some reference measure given by
\begin{equation}f_\theta(x)=\exp(\theta x - b(\theta)), \ \ \text{for} \
  \theta\in \Theta \subset \R.\label{DensityExpo}\end{equation}} Distributions
from a canonical one-parameter exponential family can be parameterized either by
their natural parameter $\theta$ or by their mean. Indeed
$\dot{b}(\theta)=\mu(\theta)$, the mean of the distribution $\nu_{\theta}$ and
$\ddot{b}(\theta)=\text{Var}[\nu_{\theta}]>0$. The mapping $\theta \mapsto
\mu(\theta)$ is strictly increasing, and the means are ordered in the same way
as the natural parameters. Exponential families include in particular Bernoulli
distributions, or Gaussian distributions with common variances (see
\cite{KLUCB:Journal} for more details about exponential families).

\begin{figure}[h]
 \centering
\includegraphics[height=5cm]{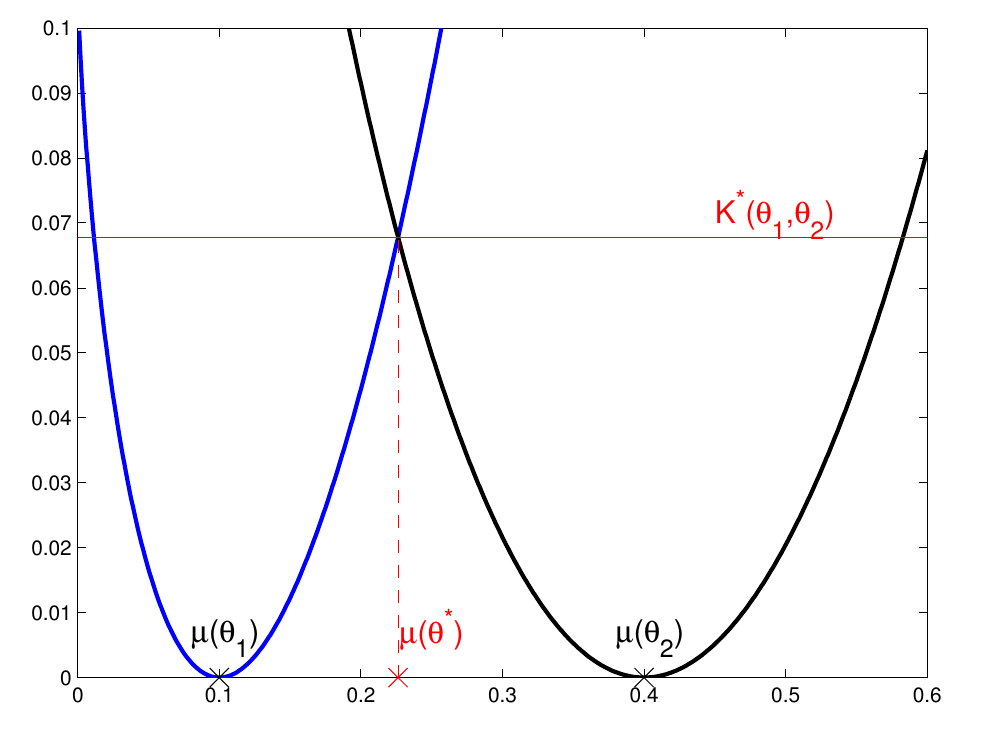} 
\caption{\label{fig:Chernoff} For Bernoulli distributions, the blue and black curves represent respectively $\text{KL}(\cB(\mu),\cB(\mu_1))$ and  $\text{KL}(\cB(\mu),\cB(\mu_2))$
as a function of $\mu$. Their intersection gives the value of the Chernoff information between $\cB(\mu_1)$ and $\cB(\mu_2)$, two distributions alternatively parameterized by their natural parameter $\theta_1$ and $\theta_2$.} 
\end{figure}

We introduce the following shorthand to denote the Kullback-Leibler divergence in
exponential families: $\Kb(\theta,\theta')=\K(\nu_{\theta},\nu_{\theta'})$ for
$(\theta,\theta')\in \Theta^2$. Combining the upper bound on the sample
complexity of the KL-LUCB algorithm obtained by \cite{COLT13} and the lower bound of Theorem
\ref{thm:GeneralBoundFC}, the complexity $\kappa_C(\nu)$ can be bounded as
\begin{equation}
  \sum_{a\in\cS_m^*}\frac{1}{\Kb(\theta_a,\theta_{[m+1]})} +
  \sum_{a\notin\cS_m^*}\frac{1}{\Kb(\theta_a,\theta_{[m]})} \leq \kappa_C(\nu) \leq 
  24 \min_{\theta \in \left[\theta_{[m+1]},\theta_{[m]}\right]} \sum_{a=1}^{K} \frac{1}{\Kb^*(\theta_a,\theta)}, \label{InfoLBUB}
\end{equation}
where $\Kb^*(\theta,\theta')$ is the Chernoff information between the
distributions $\nu_{\theta}$ and $\nu_{\theta'}$ 
(see \cite{Cover:Thomas} and \cite{COLT13} for earlier notice of the
 relevance of this quantity in the best-arm selection problem). Chernoff
 information is defined as follows and illustrated in Figure~\ref{fig:Chernoff}:
\[\Kb^*(\theta,\theta')=\Kb(\theta^*,\theta), \ \text{where} \ \theta^* \
\text{is such that} \ \ \Kb(\theta^*,\theta)=\Kb(\theta^*,\theta').\]

\section{Improved Lower Bounds for Two-Armed Bandits \label{sec:2arms}}

Two armed-bandits are of particular interest as they offer a theoretical
framework for sequential A/B Testing.  A/B Testing is a popular procedure used,
for instance, for website optimization: two versions of a web page, say A and B,
are empirically compared by being presented to users.  Each user is shown only
one version $A_t\in\{1,2\}$ and provides a real-valued index of the quality of
the page, $Z_t$, which is modeled as a sample of a probability distribution
$\nu_1$ or $\nu_2$. For example, a standard objective is to determine which
web page has the highest conversion rate (probability that a user actually
becomes a customer) by receiving binary feedback from the users. In standard
A/B Testing algorithms, the two versions are presented equally often. It is
thus of particular interest to investigate whether uniform sampling is optimal
or not.

Even for two-armed bandits, the upper and lower bounds on the complexity
$\kappa_C(\nu)$ given in (\ref{InfoLBUB}) do not match. We propose in this
section a refined lower bound on $\kappa_C(\nu)$ based on a different change of
distribution. This lower bound features a quantity reminiscent of Chernoff
information, and we will exhibit algorithms matching (or approximately
matching) this new bound in Section \ref{sec:Examples}.  Theorem
\ref{thm:2armsFC} provides a non-asymptotic lower bound on the sample
complexity $\bE_\nu[\tau]$ of any $\delta$-PAC algorithm. It also provides a
lower bound on the performance of algorithms using a uniform sampling strategy,
which will turn out to be efficient in some cases.

\begin{theorem}\label{thm:2armsFC} Let $\cM$ be an identifiable class of two-armed bandit models 
  and let $\mathbf{\nu}=(\nu_{1},\nu_{2})\in\cM$ be such that
  $\mu_1>\mu_2$. Any algorithm that is $\delta$-PAC on $\cM$ satisfies, for all
  $\delta \in )0,1]$,
  \[
  \bE_\nu[\tau]\geq \frac{1}{c_*(\nu)} \log\left(\frac{1}{2.4 \delta}\right), \ \
   \text{where} \ \ \ c_*(\nu) := \inf_{(\nu_1',\nu_2')\in\cM: \mu_1'<\mu_2'}
  \max\left\{\K(\nu_{1},\nu_{1}'),\K(\nu_{2},\nu_{2}')\right\}.
  \]
  Moreover, any $\delta$-PAC algorithm using a uniform sampling strategy
  satisfies, 
  \begin{equation}
    \bE_\nu[\tau] \geq \frac{1}{I_*(\nu)}\log \left(\frac{1}{2.4 \delta}\right), \   \text{where} 
    \  \ I_*(\nu):=\!\inf_{(\nu_1',\nu_2')\in\cM: \mu'_1 <\mu'_2}\!\frac{\K\left(\nu_1,\nu_1'\right)+\K\left(\nu_2,\nu_2'\right)}{2}.\hspace{0.5cm}
    \label{ineq:FCUnif}
  \end{equation}
\end{theorem}

Obviously, one has $I_*(\nu) \leq c_*(\nu)$. Theorem \ref{thm:2armsFC} implies in particular that $\kappa_C(\nu) \geq
1/{c_*(\nu)}$.
It is possible to give explicit expressions for the quantities $c_*(\nu)$ and
$I_*(\nu)$ for important classes of parametric bandit models that will be
considered in the next section.

The class of Gaussian bandits with known variances $\sigma_1^2$ and
$\sigma_2^2$, further considered in Section~\ref{sec:GaussianBandits}, is
\begin{equation}\cM = \{ \nu =
  \left(\norm{\mu_1}{\sigma_1^2},\norm{\mu_2}{\sigma_2^2}\right) :
  (\mu_1,\mu_2)\in\R^2, \mu_1\neq \mu_2\}.\label{set:Gaussian}\end{equation}
{For this class, 
\begin{equation}
  \label{eq:kl_gauss}
  \K(\norm{\mu_1}{\sigma_1},\norm{\mu_2}{\sigma_2}) = \frac{(\mu_1 -
  \mu_2)^2}{2\sigma_2^2} + \frac{1}{2}\left[\frac{\sigma_1^2}{\sigma_2^2} - 1 - \log
  \frac{\sigma_1^2}{\sigma_2^2}\right]  
\end{equation}
and direct computations yield}
\begin{eqnarray*}
  c_*(\nu) = \frac{(\mu_1-\mu_2)^2}{2(\sigma_1+\sigma_2)^2} \ \ \ \text{and} \ \ \ I_*(\nu)=\frac{(\mu_1 - \mu_2)^2}{4(\sigma_1^2 + \sigma_2^2)}.
\end{eqnarray*}
The observation that, when the variances are different $c_*(\nu) > I_*(\nu)$, will be shown to imply that strategies based on uniform sampling are sub-optimal {(by a factor $1 \leq 2(\sigma_1^2+\sigma_2^2)/(\sigma_1+\sigma_2)^2 \leq 2$).}

The more general class of two-armed exponential bandit models, further
considered in Section~\ref{sec:Bernoulli}, is
\[\cM = \{ \nu = (\nu_{\theta_1},\nu_{\theta_2}) :
(\theta_1,\theta_2)\in\Theta^2, \theta_1\neq \theta_2\}\] where
$\nu_{\theta_a}$ has density $f_{\theta_a}$ given by (\ref{DensityExpo}). There
\[
c_*(\nu) = \inf_{\theta\in\Theta} \max
\left(\Kb(\theta_1,\theta),\Kb(\theta_2,\theta)\right) =
\Kb_*(\theta_1,\theta_2),\] where
$\Kb_*(\theta_1,\theta_2)=\Kb(\theta_1,\theta_*)$, with $\theta_*$ is defined
by $\Kb(\theta_1,\theta_*)=\Kb(\theta_2,\theta_*)$.  This quantity is analogous to the Chernoff information $\Kb^*(\theta_1,\theta_2)$
introduced in Section \ref{sec:GeneralBoundFC} but with `reversed' roles for the arguments. $I_*(\nu)$ may also be expressed more explicitly as 
\[ I_*(\nu) =
\frac{\Kb\left(\theta_1,\overline{\theta}\right)+\Kb\left(\theta_2,
    \overline{\theta}\right)}{2}, \ \ \text{where} \ \
\mu(\overline{\theta})=\frac{\mu_1+\mu_2}{2}.\]

Appendix~\ref{sec:prop_k_star} provides further useful properties of these quantities and in particular Figure~\ref{fig:k_star} illustrates the property that for two-armed exponential bandit models, the lower bound on
$\kappa_C(\nu)$ provided by Theorem \ref{thm:2armsFC},
\begin{equation}\kappa_C(\nu) \geq
  \left(\frac{1}{\Kb_*(\theta_1,\theta_2)}\right),
  \label{eqn:Bound2arms}\end{equation} 
is indeed always tighter than the lower bound of Theorem \ref{thm:GeneralBoundFC},
\begin{equation}\hspace{2cm}\kappa_C(\nu) \geq
  \left(\frac{1}{\Kb(\theta_1,\theta_2)} +
    \frac{1}{\Kb(\theta_2,\theta_1)}\right)\label{eqn:BoundGene}.\end{equation}

\begin{figure}[h]
  \includegraphics[angle=-90,width=\textwidth]{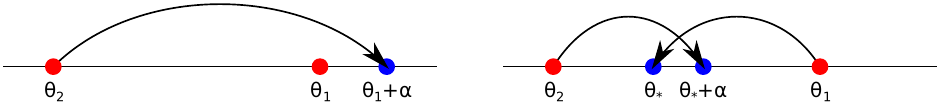}
  \caption{\label{fig:IllustrChangeDist}Alternative bandit models considered to
    obtain the lower bounds of Theorem \ref{thm:GeneralBoundFC} (left) and
    Theorem \ref{thm:2armsFC} (right).}
\end{figure}

Interestingly, the changes of distribution used to derive the two results are
not the same.  On the one hand, for inequality~(\ref{eqn:BoundGene}), the
changes of distribution involved modify a single arm at a time: one of the arms
is moved just below (or just above) the other (see
Figure~\ref{fig:IllustrChangeDist}, left).  This is the idea also used, for
example, to obtain the lower bound of \cite{LaiRobbins85bandits} on the
cumulative regret.  On the other hand, for inequality~(\ref{eqn:Bound2arms}),
both arms are modified at the same time: they are moved close to the common
intermediate value $\theta_*$ but with a reversed ordering
(see Figure~\ref{fig:IllustrChangeDist}, right).

We now give the proof of Theorem \ref{thm:2armsFC}, in order to show how easily
it follows from Lemma~\ref{lem:Cornerstone}.

\medskip

\emph{Proof of Theorem \ref{thm:2armsFC}.} Without loss of generality, one may assume that the bandit model
$\nu=(\nu_1,\nu_2)$ is such that the best arm is $a^*=1$. Consider any
alternative bandit model $\nu'=(\nu_1',\nu_2')$ in which $a^*=2$. Let $\cE$ be
the event $\cE=(\hat{\cS}_1=1)$, which belongs to $\cF_\tau.$

Let $\cA=((A_t),\tau,\hat{S}_1)$ be a $\delta$-PAC algorithm: by assumptions,
$\bP_\nu(\cE)\geq 1-\delta$ and $\bP_{\nu'}(\cE) \leq \delta$.  Applying Lemma
\ref{lem:Cornerstone} (with the stopping time $\tau$) and using again the
monotonicity properties of $d(x,y)$ and inequality~\eqref{TrickKL}
\begin{equation}
  \bE_\nu[N_1]\K(\nu_{1},\nu_{1}') + \bE_\nu[N_2]\K(\nu_{2},\nu_{2}') \geq \log (1/(2.4\delta)).\label{GeneralPart} 
\end{equation}
Using moreover that $\tau = N_1 + N_2$, one has
\begin{equation}\bE_\nu[\tau] \geq
\frac{\log\left(\frac{1}{2.4 \delta}\right)}{\max_{a=1,2}\K(\nu_{a},\nu_{a}')}.\label{ThisIneq}\end{equation}
The result follows by optimizing over the possible model $\nu'$ satisfying
$\mu_1'<\mu_2'$ to make the right hand side of the inequality as large as
possible. More precisely, for every $\alpha>0$, from the definition of $c_*(\nu)$, there exists $\nu'_\alpha=(\nu_1',\nu_2')$ for which \[\max_{a=1,2}\K(\nu_{a},\nu_{a}')< c_*(\nu) + \alpha.\]
Inequality~\eqref{ThisIneq} for the particular choice $\nu'=\nu'_\alpha$ yields 
$\bE_\nu[\tau] \geq (c_*(\nu)+\alpha)^{-1}\log(1/(2.4\delta))$, and the first statement of Theorem~\ref{thm:2armsFC} follows by letting $\alpha$ go to zero. In the particular case of exponential bandit models, the alternative model  consists in choosing $\nu_1'=\nu_{\theta_*}$ and $\nu_2'=\nu_{\theta_* + \epsilon}$ for some $\epsilon$, as illustrated on Figure~\ref{fig:IllustrChangeDist}, so that $\max_{a=1,2}\K(\nu_{a},\nu_{a}')$ is of order $\Kb_*(\theta_1,\theta_2).$

When $\cA$ uses uniform sampling, using the fact that $\bE_\nu[N_1]=\bE[N_2]=\bE[\tau]/2$ in Equation (\ref{GeneralPart}) similarly gives the second statement of Theorem \ref{thm:2armsFC}.

\section{Matching Algorithms for Two-Armed Bandits\label{sec:Examples}}
{For specific instances of two-armed bandit models, we now present algorithms
with performance guarantees that closely match the lower bounds of Theorem
\ref{thm:2armsFC}. For Gaussian bandits with known (and possibly different)
variances, we describe in Section \ref{sec:GaussianBandits} an algorithm termed
$\alpha$-Elimination that is optimal and thus makes it possible to determine
the complexity $\kappa_C(\nu)$. For Bernoulli bandit models, we present in
Section \ref{sec:Bernoulli} the SGLRT algorithm that uses uniform sampling and
is close to optimal.}

\subsection{Gaussian Bandit Models \label{sec:GaussianBandits}} 
We focus here on the class of two-armed Gaussian bandit models with known
variances presented in (\ref{set:Gaussian}), where $\sigma_1$ and $\sigma_2$
are fixed. We prove that
\[\kappa_C(\nu)=\frac{2(\sigma_1+\sigma_2)^2}{(\mu_1-\mu_2)^2}\]
by exhibiting a strategy that reaches the performance bound of Theorem
\ref{thm:2armsFC}.  This strategy uses non-uniform sampling in case where
$\sigma_1$ and $\sigma_2$ differ. When $\sigma_1 = \sigma_2$, we provide in
Theorem~\ref{thm:PACFC} an improved stopping rule that is $\delta$-PAC and
results in a significant reduction of the expected number of samples used.

The $\alpha$-Elimination algorithm introduced in this Section can also be used
in more general two-armed bandit models, where the distribution $\nu_a$ is
$\sigma_a^2$-subgaussian. This means that the probability distribution $\nu_a$
satisfies
\[\forall \lambda \in \R, \ \ \bE_{X\sim \nu_a}\left[e^{\lambda X}\right]\leq
\frac{\lambda^2\sigma_a^2}{2}.\] This covers in particular the cases of bounded
distributions with support in $[0,1]$ (that are $1/4$-subgaussian).  In these
more general cases, the algorithm enjoys the same theoretical properties: it is
$\delta$-PAC and its sample complexity is bounded as in Theorem
\ref{thm:MatchingFC} below. 

\subsubsection{Equal Variances \label{subsec:equalvar}}

We start with the simpler case $\sigma_1=\sigma_2=\sigma$. Thus, the quantity
$I_*(\nu)$ introduced in Theorem \ref{thm:2armsFC} coincides with $c_*(\nu)$,
which suggests that uniform sampling could be optimal.  A uniform sampling
strategy equivalently collects paired samples $(X_s,Y_s)$ from both
arms. The difference $X_s - Y_s$ is normally distributed with mean
$\mu=\mu_1-\mu_2$ and a $\delta$-PAC algorithm is equivalent to a sequential
test of $H_0:(\mu<0)$ versus $H_1:(\mu>0)$ such that both type I and type II
error probabilities are bounded by $\delta$. \cite{Robbins70LIL} proposes the
stopping rule
\begin{equation}\label{Robbins}
  \tau = \inf\Bigg\{ t\in 2\N^*: \Big|\sum_{s=1}^{t/2} (X_s - Y_s)\Big| > \sqrt{2\sigma^2t\beta(t,\delta)}\Bigg\}, 
  \text{with} \ \beta(t,\delta)=\frac{t+1}{t}\log\left(\frac{t+1}{2\delta}\right).
\end{equation}
The recommendation rule chooses the empirically best arm at time $\tau$. This
procedure can be seen as an \textit{elimination strategy}, in the sense of
\cite{Jennison:al84}. The authors of this paper derive a lower bound on the
sample complexity of any $\delta$-PAC \textit{elimination} strategy (whereas
our lower bound applies to \textit{any} $\delta$-PAC algorithm) which is
matched by Robbins' algorithm: the above stopping rule $\tau$ satisfies
\[\lim_{\delta \rightarrow
  0}\frac{\bE_\nu[\tau]}{\log({1}/{\delta})} =
\frac{8\sigma^2}{(\mu_1-\mu_2)^2}.\] This value coincide with the lower bound
on $\kappa_C(\nu)$ of Theorem \ref{thm:2armsFC} in the case of two-armed
Gaussian distributions with similar known variance $\sigma^2$. This proves that
in this case, Robbins' rule (\ref{Robbins}) is not only optimal among the class
of elimination strategies, but also among the class of $\delta$-PAC algorithm.

Any $\delta$-PAC elimination strategy that uses a threshold function (or
\textit{exploration rate}) $\beta(t,\delta)$ smaller than Robbins' also matches
our asymptotic lower bound, while stopping earlier than the latter.  From a
practical point of view, it is therefore interesting to exhibit smaller
exploration rates that preserve the $\delta$-PAC property. The failure
probability of such an algorithm is upper bounded, for example when
$\mu_1<\mu_2$, by
\begin{equation}\label{PACExplain}
  \bP_\nu\left(\exists k\in\N : \sum_{s=1}^{k} \frac{X_s - Y_s - (\mu_1-\mu_2)}{\sqrt{2\sigma^2}}
    > \sqrt{2k\beta(2k,\delta)}\right) =  \bP\left(\exists k\in\N : S_k > \sqrt{2k\beta(2k,\delta)}\right)
\end{equation}
where $S_k$ is a sum of $k$ i.i.d. variables of distribution
$\norm{0}{1}$. \cite{Robbins70LIL} obtains a non-explicit confidence region of
risk at most $\delta$ by choosing
$\beta(2k,\delta)=\log\left({\log(k)}/{\delta}\right) + o(\log\log(k))$. The
dependency in $k$ is in some sense optimal, because the Law of Iterated
Logarithm (LIL) states that
$\limsup_{k\rightarrow\infty}{S_k}/\sqrt{2k\log\log(k)} =1$ almost
surely. In this paper, we propose a new deviation inequality for a martingale with sub-Gaussian increments, stated as Lemma~\ref{thm:subgaussian}, 
that permits to build an explicit confidence region reminiscent of the LIL. A related result was recently derived independently by \cite{Jamiesonal14LILUCB}. 

\begin{lemma}\label{thm:subgaussian}
Let $\zeta(u) = \sum_{k\geq 1} k^{-u}$.  Let $X_1, X_2,\dots$ be independent random variables such that, for all $\lambda\in\R$, $\phi(\lambda):=\log\bE[\exp(\lambda X_1)] \leq \lambda^2\sigma^2/2$.  For every positive integer $t$ let $S_t = X_1 + \dots + X_t$.  Then, for all $\eta>1$ and $x\geq \frac{8}{(e-1)^2}$,
\[\bP\left(\exists t\in \N^* : {S_t} > \sqrt{2\sigma^2t(x + \eta\log\log(et))}\right) \leq \sqrt{e}\, \zeta\left(\eta \left(1-\frac{1}{2x}\right)\right)\left(\frac{\sqrt{x}}{2\sqrt{2}}+1\right)^{\eta}\exp(-x).\]
\end{lemma}

Lemma~\ref{thm:subgaussian} allows to prove Theorem~\ref{thm:PACFC} below, as detailed in Appendix~\ref{proof:DevIneq}, where we also provide a proof of Lemma~\ref{thm:subgaussian}.

\begin{theorem} \label{thm:PACFC} For $\delta \leq 0.1$, with
  \begin{equation}\label{choiceBetaBetter}\beta(t,\delta)=\log ({1}/{\delta})
    + 3\log\log ({1}/{\delta}) +    (3/2)\log(\log(et/2)),\end{equation}
  the elimination strategy is $\delta$-PAC. 
\end{theorem}

We refer to Section \ref{sec:Experiments} for numerical simulations that illustrate the significant savings 
(in the average number of samples needed to reach a decision) resulting from the use of 
the less conservative exploration rate allowed by Theorem~\ref{thm:PACFC}.

\subsubsection{Mismatched Variances\label{subsec:General}}

In the case where $\sigma_1\neq \sigma_2$, we rely on the $\alpha$-Elimination
strategy, described in Algorithm~\ref{AlgoBox:Elimination} below.  For $a=1,2$,
$\hat{\mu}_{a}(t)$ denotes the empirical mean of the samples gathered from arm
$a$ up to time $t$.  The algorithm is based on a non-uniform sampling strategy
governed by the parameter $\alpha\in(0,1)$, that maintains the proportion of draws of arm 1 
close to $\alpha$. At the end of every round $t$, $N_1(t)=\lceil \alpha t\rceil$, $N_2(t)=t - \lceil \alpha
t\rceil$ and $\hat{\mu}_1(t)-\hat{\mu}_2(t) \sim \norm{\mu_1 -
  \mu_2}{\sigma_t^2(\alpha)}$ (where $\sigma_t^2(\alpha)$ is defined at line 6
of Algorithm \ref{AlgoBox:Elimination}). The sampling schedule used here is
thus deterministic.

\begin{algorithm}[hbt]
  \caption{$\alpha$-Elimination\label{AlgoBox:Elimination}}
  \begin{algorithmic}[1]
    \REQUIRE Exploration function $\beta(t,\delta)$, parameter $\alpha$.
    \STATE \textit{Initialization}: $\hat{\mu}_1(0)=\hat{\mu}_2(0)=0$,
    $\sigma^2_{0}(\alpha)=1$, $t=0$ \WHILE {$|\hat{\mu}_1(t) - \hat{\mu}_2(t)|
      \leq \sqrt{2\sigma^2_{t}(\alpha)\beta(t,\delta)}$} \STATE $t \leftarrow
    t+1$.  \STATE If $\lceil\alpha t \rceil = \lceil \alpha(t-1) \rceil$,
    $A_{t}\leftarrow 2$, else $A_t\leftarrow 1$ \STATE Observe $Z_t\sim
    \nu_{A_t}$ and compute the empirical means $\hat{\mu}_1(t)$ and
    $\hat{\mu}_2(t)$ \STATE Compute $\sigma_t^2(\alpha)=\sigma_1^2/\lceil\alpha
    t\rceil + \sigma_2^2/(t-\lceil \alpha t \rceil)$
    \ENDWHILE
    \RETURN $\argmax{a=1,2} \ \hat{\mu}_a(t)$
  \end{algorithmic}
\end{algorithm}

Theorem \ref{thm:MatchingFC} shows that an optimal allocation of samples between the two arms consists in maintaining the proportion of draws of arm 1 close to $\sigma_1/(\sigma_1 +
\sigma_2)$ (which is also the case in the fixed-budget setting, see Section~\ref{CompGauss}). Indeed, for $\alpha=\sigma_1/(\sigma_1 +
\sigma_2)$, the $\alpha$-elimination algorithm is
$\delta$-PAC with a suitable exploration rate and (almost) matches the lower bound on $\bE_\nu[\tau]$, at least
asymptotically when $\delta \rightarrow 0$. Its proof can be found in Appendix
\ref{proof:MatchingFC}.

\begin{theorem}\label{thm:MatchingFC} If $\alpha=\sigma_1/(\sigma_1 + \sigma_2)$, the $\alpha$-elimination strategy using the exploration rate  
  $\beta(t,\delta)=\log\frac{t}{\delta} + 2 \log\log(6t)$ is $\delta$-PAC on
  $\cM$ and satisfies, for every $\nu\in\cM$, for every $\epsilon>0$,
$$\bE_\nu[\tau] \leq (1+\epsilon)\frac{2(\sigma_1 + \sigma_2)^2}{(\mu_1-\mu_2)^2}\log\left(\frac{1}{\delta}\right) + \underset{\delta \rightarrow 0}{o_{\epsilon}}\left(\log\left(\frac{1}{\delta}\right)\right).$$
\end{theorem}

\begin{remark} When $\sigma_1=\sigma_2$, $1/2$-elimination reduces, up to
  rounding effects, to the elimination procedure described in
  Section~\ref{subsec:equalvar}, for which Theorem \ref{thm:PACFC} suggests an
  exploration rate of order $\log(\log(t)/\delta)$. As the feasibility of this
  exploration rate when $\sigma_1\neq\sigma_2$ is yet to be established, we
  focus on Gaussian bandits with equal variances in the numerical experiments
  of Section~\ref{sec:Experiments}.
\end{remark}

\subsection{Bernoulli Bandit Models \label{sec:Bernoulli}}

We consider in this section the class of Bernoulli bandit models
\[\cM = \{ \nu=\left(\cB(\mu_1),\cB(\mu_2)\right) : (\mu_1,\mu_2)\in (0;1)^2,
\mu_1 \neq \mu_2 \},\] where each arm can be alternatively parameterized by the
natural parameter of the exponential family, $\theta_a =
\log({\mu_a}/{(1-\mu_a)})$. Observing that in this particular case little can be
gained by departing from uniform sampling, {we consider the SGLRT algorithm (to be defined below) that uses uniform sampling together with a stopping rule that is not based on the mere difference of the empirical means.}

{For Bernoulli bandit models, the quantities $I_*(\nu)$ and $c_*(\nu)$ introduced in
Theorem \ref{thm:2armsFC} happen to be practically very close (see Figure
\ref{fig:CompareComplexities} in Section \ref{sec:FixedBudget} below). There is
thus a strong incentive to use uniform sampling and in the rest of
this section we consider algorithms that aim at matching the bound
(\ref{ineq:FCUnif}) of Theorem \ref{thm:2armsFC}---that is, $\bE_\nu[\tau]
\leq \log(1/\delta)/I_*(\nu)$, at least for small values of $\delta$---, which
provides an upper bound on $\kappa_C(\nu)$ that is very close to $1/c_*(\nu)$. 
For simplicity, as $I_*(\nu)$ is here a function of the 
means of the arms only, we will denote $I_*(\nu)$ by $I_*(\mu_1,\mu_2)$.}

When the arms are sampled uniformly, finding an algorithm that matches the bound
of (\ref{ineq:FCUnif}) boils down to determining a proper stopping rule.
In all the algorithms studied so
far, the stopping rule was based on the difference of the empirical means of
the arms. For Bernoulli arms the 1/2-Elimination procedure described in Algorithm~\ref{AlgoBox:Elimination} can be used, as each
distribution $\nu_a$ is bounded and therefore 1/4-subgaussian. More precisely,
with $\beta(t,\delta)$ as in Theorem \ref{thm:PACFC}, the algorithm stopping at
the first time $t$ such that
\[\hat{\mu}_{1}(t) - \hat{\mu}_{2}(t) > \sqrt{{2\beta(t,\delta)}/{t}}\]
has its sample complexity bounded by ${2}/{(\mu_1-\mu_2)^2}\log({1}/{\delta}) +
o \left(\log({1}/{\delta})\right)$. Yet, Pinsker's inequality implies that
$I_*(\mu_1,\mu_2)>(\mu_1-\mu_2)^2/2$ and this algorithm is thus not optimal
with respect to the bound (\ref{ineq:FCUnif}) of Theorem \ref{thm:2armsFC}.
The approximation $I_*(\mu_1,\mu_2) = (\mu_1-\mu_2)^2/(8\mu_1(1-\mu_1)) +
o\left((\mu_1-\mu_2)^2\right)$ suggests that the loss with respect to the
optimal error exponent is particularly significant when both means are close to
0 or 1.

\begin{algorithm}[b]
  \caption{Sequential Generalized Likelihood Ratio Test
    (SGLRT)\label{AlgoBox:SGLRT}}
  \begin{algorithmic}[1]
    \REQUIRE Exploration function $\beta(t,\delta)$.
    \STATE \textit{Initialization}: $\hat{\mu}_1(0)=\hat{\mu}_2(0)=0$. $t=0$.
    \WHILE {($tI_*(\hat{\mu}_1(t),\hat{\mu}_2(t))\leq
      \beta(t,\delta))\bigcup(t=1 \ (mod. \ 2))$} \STATE $t=t+1$. $A_{t}=t \ (mod. \ 2)$.  \STATE
    Observe $Z_t\sim \nu_{A_t}$ and compute the empirical means
    $\hat{\mu}_1(t)$ and $\hat{\mu}_2(t)$.
    \ENDWHILE
    \RETURN $a=\argmax{a=1,2} \ \hat{\mu}_a(t)$.
  \end{algorithmic}
\end{algorithm}

To circumvent this drawback, we propose the SGLRT (for Sequential
Generalized Likelihood Ratio Test) stopping rule, described in Algorithm
\ref{AlgoBox:SGLRT}. The appearance of $I_*$ in the stopping criterion of Algorithm
\ref{AlgoBox:SGLRT} is a consequence of the observation that it is related to the generalized likelihood ratio statistic for testing the equality of two Bernoulli proportions. To test $H_0:(\mu_1=\mu_2)$ against $H_1:(\mu_1\neq\mu_2)$ based on $t/2$ paired samples of the arms $W_s=(X_s,Y_s)$, the Generalized Likelihood Ratio Test (GLRT) rejects $H_0$ when
\[\frac{\max_{\mu_1,\mu_2 : \mu_1=\mu_2} L(W_1,\dots,W_{t/2};
  \mu_1,\mu_2)}{\max_{\mu_1,\mu_2} L(W_1,\dots,W_{t/2} ; \mu_1,\mu_2)} <
z_\delta,\] where $L(W_1,\dots,W_{t/2} ; \mu_1,\mu_2)$ denotes the likelihood of the observations given parameters $\mu_1$ and $\mu_2$. It can be checked that the ratio that appears in the last display is equal to 
$\exp(-tI_*(\hat{\mu}_{1,t/2},\hat{\mu}_{2,t/2}))$. This equality is a consequence of the rewriting
\begin{eqnarray*}
  I_*(x,y)&=&H\left(\frac{x+y}{2}\right)-\frac{1}{2}\left[H\left({x}\right)+H\left({y}\right)\right],
\end{eqnarray*}
where $H(x)=-x\log(x)-(1-x)\log(1-x)$ denotes the binary entropy function. Hence, Algorithm
(\ref{AlgoBox:SGLRT}) can be interpreted as a sequential version of the GLRT
with (varying) threshold $z_{t,\delta}=\exp(-\beta(t,\delta))$.

\medskip

\emph{Elements of analysis of the SGLRT.} The SGLRT algorithm is also related to the KL-LUCB algorithm of
\cite{COLT13}. A closer examination of the KL-LUCB stopping criterion reveals
that, in the specific case of two-armed bandits, it is equivalent to stopping
when $t\K_*(\cB(\hat{\mu}_{1}(t)),\cB(\hat{\mu}_{2}(t)))$ gets larger than some
threshold. We also mentioned the fact that $\K_*(\cB(x),\cB(y))$ and $I_*(x,y)$ are 
very close (see Figure \ref{fig:CompareComplexities}). Using results from
\cite{COLT13}, one can thus prove (see Appendix \ref{sec:DetailsBernoulli}) the following
lemma.

\begin{lemma}\label{lem:OptBernoulliKL} With the exploration rate
  \[\beta(t,\delta)=2\log\left(\frac{t(\log(3t))^2}{\delta}\right)\]
  the SGLRT algorithm is $\delta$-PAC.
\end{lemma}

For this exploration rate, we were able to obtain the following asymptotic guarantee on the stopping time $\tau$ of Algorithm \ref{AlgoBox:SGLRT}:
\[\forall\epsilon>0, \ \ \limsup_{\delta\rightarrow 0}
\frac{\tau}{\log(1/\delta)} \leq \frac{2(1+\epsilon)}{I_*(\mu_1,\mu_2)} \ \
a.s. \] (see Lemma \ref{lem:OptBernoulliKL2} in Appendix
\ref{sec:DetailsBernoulli} for the proof of this result). By analogy with the
result of Theorem \ref{thm:PACFC} we conjecture that the analysis of
\cite{COLT13}---on which the result of Lemma \ref{lem:OptBernoulliKL} is
based---is too conservative and that the use of an exploration rate of order
$\log(\log(t)/\delta)$ should also lead to a $\delta$-PAC algorithm. 
This conjecture is supported by the numerical experiments reported in Section
\ref{sec:Experiments} below. Besides, for this choice of  exploration rate, Lemma~\ref{lem:OptBernoulliKL2} also shows that 
\[\forall\epsilon>0, \ \ \limsup_{\delta\rightarrow 0}
\frac{\tau}{\log(1/\delta)} \leq \frac{(1+\epsilon)}{I_*(\mu_1,\mu_2)} \ \
a.s. . \]

\section{The Fixed-Budget Setting \label{sec:FixedBudget}}

In this section, we focus on the fixed-budget setting and we provide new upper
and lower bounds on the complexity term $\kappa_B(\nu)$.

For two-armed bandits, we obtain in Theorem \ref{thm:2armsFB} lower bounds
analogous to those of Theorem \ref{thm:2armsFC} in the fixed-confidence
setting. We present matching algorithms for Gaussian and Bernoulli
bandits. This allows for a comparison between the fixed-budget and
fixed-confidence setting in these specific cases. More specifically, we show
that $\kappa_B(\nu)=\kappa_C(\nu)$ for Gaussian bandit models, whereas
$\kappa_C(\nu)>\kappa_B(\nu)$ for Bernoulli bandit models.

When $K>2$ and $m\geq 1$, we present a first step towards obtaining more
general results, by providing lower bounds on the probability of error
$p_t(\nu)$ for Gaussian bandits with equal variances.

\subsection{Comparison of the Complexities for Two-Armed Bandits \label{CompGauss}}

We present here an asymptotic lower bound on $p_t(\nu)$ that directly yields a
lower bound on $\kappa_B(\nu)$. Moreover, we provide a lower bound on the
failure probability of consistent algorithms using uniform sampling. The proof
of Theorem \ref{thm:2armsFB} bears similarities with that of Theorem
\ref{thm:2armsFC}, and we provide it in Appendix \ref{proof:2armsFB}. However,
it is important to note that the informational quantities $c^*(\nu)$ and
$I^*(\nu)$ defined in Theorem \ref{thm:2armsFB} are in general different from
the quantities $c_*(\nu)$ and $I_*(\nu)$ previously defined for the
fixed-confidence setting (see Theorem
\ref{thm:2armsFC}). Appendix~\ref{sec:prop_k_star} contains a few additional
elements of comparison between these quantities in the case of one-parameter
exponential families of distributions.

\begin{theorem}\label{thm:2armsFB}Let $\mathbf{\nu}=(\nu_{1},\nu_{2})$ be a two-armed bandit model such that $\mu_1>\mu_2$. 
  In the fixed-budget setting, any consistent algorithm satisfies
  \[
  \limsup_{t\rightarrow \infty}-\frac{1}{t} \log p_t(\nu) \leq c^*(\nu),\ \ \
  \text{where} \ \ \ c^*(\nu) := \inf_{(\nu_1',\nu_2')\in\cM: \mu'_1 <\mu'_2}
  \max\left\{\K(\nu_1',\nu_{1}),\K(\nu_2',\nu_{2})\right\}.
  \]
  Moreover, any consistent algorithm using a uniform sampling strategy
  satisfies
  \begin{equation}\limsup_{t\rightarrow\infty} -\frac{1}{t}\log p_t(\nu) \leq
    I^*(\nu), \ \ \ \text{where} \ \ \
    I^*(\nu):=\inf_{(\nu_1',\nu_2')\in\cM: \mu'_1 <\mu'_2}\frac{\K\left(\nu_1',\nu_1\right)+\K\left(\nu_2',\nu_2\right)}{2}.\hspace{0.5cm}
    \label{BoundUnifFB}
  \end{equation}
\end{theorem}

\emph{Gaussian distributions.}  As \hspace{0.1cm} the \hspace{0.1cm} Kullback-Leibler \hspace{0.1cm} divergence \hspace{0.1cm} between 
two \hspace{0.1cm} Gaussian distributions---\eqref{eq:kl_gauss}---is symmetric with respect to the means when the variances are held fixed, it holds that $c^*(\nu)=c_*(\nu)$.  To find a matching algorithm, we introduce the
simple family of \textit{static strategies} that draw $n_1$ samples from arm 1
followed by $n_2=t-n_1$ samples of arm 2, and then choose arm 1 if
$\hat{\mu}_{1,n_1}>\hat{\mu}_{2,n_2}$, where $\hat{\mu}_{i,n_i}$ denotes the
empirical mean of the $n_i$ samples from arm $i$. Assume for instance that
$\mu_1>\mu_2$. Since $\hat{\mu}_{1,n_1} - \hat{\mu}_{2,n_2}-\mu_1 + \mu_2 \sim
\norm{0}{{\sigma_1^2}/{n_1}+{\sigma_2^2}/{n_2}}$, the probability of error of
such a strategy is upper bounded by
\begin{eqnarray*}
  \bP\left(\hat{\mu}_{1,n_1} < \hat{\mu}_{2,n_2}\right) & 
  \leq & 
  \exp\left(-\left(\frac{\sigma_1^2}{n_1}+\frac{\sigma_2^2}{n_2}\right)^{-1}\frac{(\mu_1-\mu_2)^2}{2}\right). 
\end{eqnarray*}
The right hand side is minimized when $n_1/(n_1+n_2) = {\sigma_1}/{(\sigma_1 +
  \sigma_2)}$, and the static strategy drawing $n_1 = \left\lceil\sigma_1
  t/(\sigma_1 + \sigma_2) \right\rceil$ times arm 1 is such that
\[ \liminf_{t \rightarrow \infty} -\frac{1}{t}\log p_t(\nu) \geq \frac{(\mu_1 -
  \mu_2)^2}{2(\sigma_1 + \sigma_2)^2}=c^*(\nu)\;.\] This shows in
particular that for Gaussian distributions the two complexities are equal:
\[\kappa_B(\nu)=\kappa_C(\nu)= \frac{2(\sigma_1 + \sigma_2)^2}{(\mu_1 -
  \mu_2)^2}.\]

\emph{Exponential families.} For exponential family bandit models, it can be
observed that
\[
c^*(\nu) = \inf_{\theta\in\Theta} \max
\left(\Kb(\theta,\theta_1),\Kb(\theta,\theta_2)\right) =
\Kb^*(\theta_1,\theta_2),\] where $\Kb^*(\theta_1,\theta_2)$ is the Chernoff
information between the distributions $\nu_{\theta_1}$ and $\nu_{\theta_2}$. We
recall that $\Kb^*(\theta_1,\theta_2)=\Kb(\theta^*,\theta_1)$, where $\theta^*$
is defined by $\Kb(\theta^*,\theta_1)=\Kb(\theta^*,\theta_2)$. Moreover, one
has
\[
I^*(\nu) =
\frac{\Kb\left(\frac{\theta_1+\theta_2}{2},\theta_1\right)+\Kb\left(\frac{\theta_1+\theta_2}{2},
    \theta_2\right)}{2}.
\]

In particular, the quantity $c^*(\nu)=\Kb^*(\theta_1,\theta_2)$ does not always
coincide with the quantity $c_*(\nu)=\Kb_*(\theta_1,\theta_2)$ defined in
Theorem \ref{thm:2armsFC}.  More precisely, $c_*(\nu)$ and $c^*(\nu)$ are equal
when the log-partition function $b(\theta)$ is (Fenchel) self-conjugate, which
is the case for Gaussian and exponential variables (see Appendix~\ref{sec:prop_k_star}). However, for Bernoulli
distributions, it can be checked that $c^*(\nu) > c_*(\nu)$. By exhibiting a
matching strategy in the fixed-budget setting (Theorem
\ref{prop:Bernoulli1}), we show that this implies that
$\kappa_C(\nu)>\kappa_B(\nu)$ in the Bernoulli case (Theorem
\ref{prop:Bernoulli2}). We also show that in this case, only little can be
gained by departing from uniform sampling.
  
\begin{theorem}\label{prop:Bernoulli1} Consider a two-armed exponential bandit model and $\alpha(\theta_1,\theta_2)$ be
  defined by
  \[\alpha(\theta_1,\theta_2)=\frac{\theta^*-\theta_1}{\theta_2-\theta_1} \ \ \
  \text{where} \ \ \Kb(\theta^*,\theta_1)=\Kb(\theta^*,\theta_2).\] For all
  $t$, the static strategy that allocates $\left\lceil\alpha(\theta_1,\theta_2)
    t\right\rceil$ samples to arm
  1, and recommends the empirical best arm, satisfies $p_t(\nu) \leq
  \exp(-tK^*(\theta_1,\theta_2))$.
\end{theorem}

Theorem \ref{prop:Bernoulli1}, whose proof can be found in Appendix
\ref{proof:ConcExp}, shows in particular that for every exponential family bandit
model there exists a consistent static strategy such that
\[\liminf_{t\rightarrow \infty} - \frac{1}{t}\log p_t \geq
\Kb^*(\theta_1,\theta_2), \ \ \ \text{and hence that} \ \
\kappa_B(\nu)=\frac{1}{\Kb^*(\theta_1,\theta_2)}.\]

By combining this
observation with Theorem \ref{thm:2armsFC} and the fact that, $\Kb_*(\theta_1,\theta_2)<\Kb^*(\theta_1,\theta_2)$ for Bernoulli distributions, one obtains
the following inequality.

\begin{theorem}\label{prop:Bernoulli2}
  For two-armed Bernoulli bandit models, $\kappa_C(\nu)>\kappa_B(\nu)$.
\end{theorem}

Note that we have determined the complexity of the fixed-budget
setting by exhibiting an algorithm (leading to an upper bound on $\kappa_B$) that is
of limited practical interest for Bernoulli bandit models.
Indeed, the optimal static strategy defined in Theorem \ref{prop:Bernoulli1} requires the
knowledge of the quantity $\alpha(\theta_1,\theta_2)$, that depends on the unknown means of the arms.  So far, it is not known whether there exists
a \textit{universal} strategy, that would satisfy $p_t(\nu)\leq
\exp(-\Kb^*(\theta_1,\theta_2)t)$ on \textit{every} Bernoulli bandit model.

\begin{figure}[h]
  \centering
  \includegraphics[height=5cm]{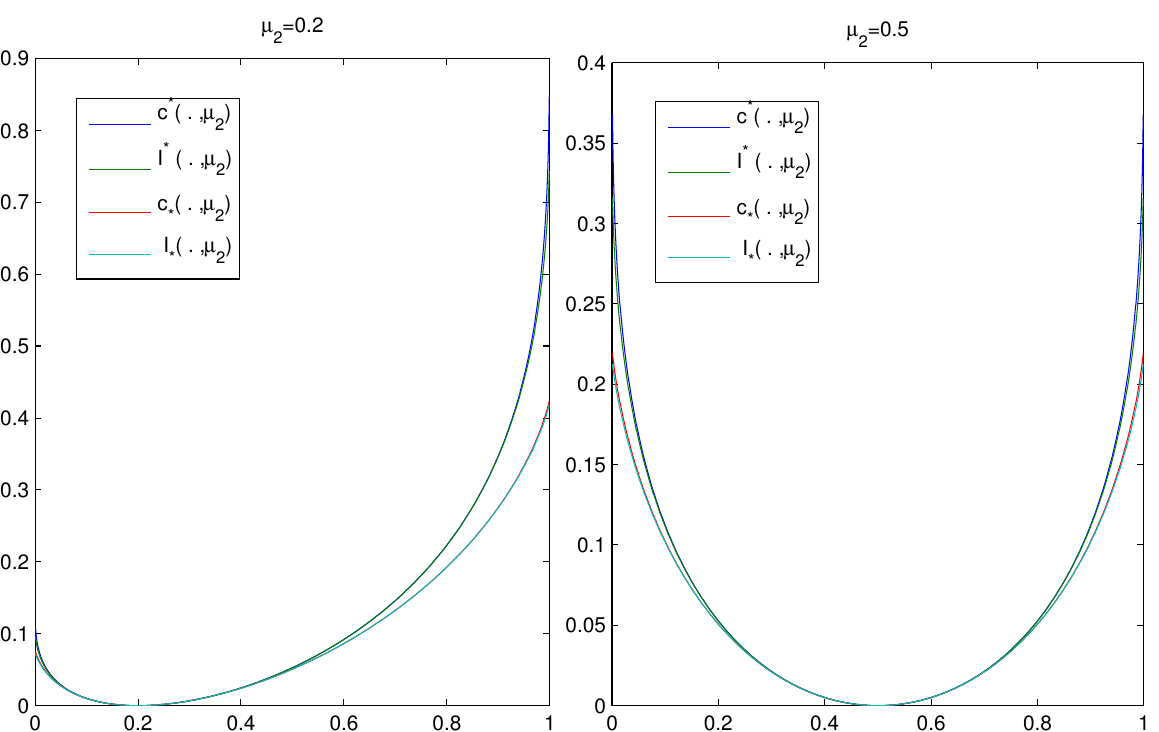}
  \caption{\label{fig:CompareComplexities} Comparison of different
    informational quantities for Bernoulli bandit models.}
\end{figure}

However, Lemma \ref{lem:ConcExp} shows that the strategy that uses uniform
sampling and recommends the empirical best-arm satisfies $p_t(\nu)\leq
\exp(-I^*(\nu)t)$, and matches the bound (\ref{BoundUnifFB}) of Theorem
\ref{thm:2armsFB} (see Remark \ref{rem:Match} in Appendix \ref{proof:ConcExp}).
The fact that, just as in the fixed-confidence setting $I^*(\nu)$ is very close
to $c^*(\nu)$ shows that the problem-dependent optimal strategy described above
can be approximated by a very simple, universal algorithm that samples the arms
uniformly.  Figure \ref{fig:CompareComplexities} represents the different
informational functions $c_*,I_*,c^*$ and $I^*$ when the mean $\mu_1$ varies,
for two fixed values of $\mu_2$. {It can be observed that $c^*(\nu)$ and
$c_*(\nu)$ are almost indistinguishable from $I^*(\nu)$ and $I_*(\nu)$, respectively, while there is a gap between $c^*(\nu)$ and $c_*(\nu)$.}

\subsection{Lower Bound on $p_t(\nu)$ in More General Cases\label{sec:FirstBoundFB}}

Theorem~\ref{thm:2armsFB} provides a direct counterpart to Theorem~\ref{thm:2armsFC}, allowing for a complete comparison between the fixed confidence and fixed budget settings in the case of two-armed bandits. However, we were not able to obtain a general lower bound for $K$-armed bandit that would be directly comparable to that of Theorem~\ref{thm:GeneralBoundFC} in the fixed budget setting. Using Lemma \ref{lem:CornerVariant} stated below (a variant of Lemma
\ref{lem:Cornerstone} proved in Appendix \ref{proof:CornerVariant}), we were nonetheless
able to derive tighter, non-asymptotic, lower bounds on $p_t(\nu)$ in
the particular case of Gaussian bandit models with equal known variance, $\cM_m=\{\nu=(\nu_1,\dots,\nu_K) : \nu_a=\norm{\mu_a}{\sigma^2} , \mu_a\in \R
, \mu_{[m]}\neq\mu_{[m+1]}\}$.

\begin{lemma}\label{lem:CornerVariant}
  Let $\nu$ and $\nu'$ be two bandit models such that $\mathcal{S}_m^*(\nu)\neq
  \mathcal{S}_m^{*}(\nu')$. Then
  \[
  \max\left(\bP_\nu(\cS \neq \cS_m^*(\nu)),\bP_{\nu'}(\cS \neq
    \cS_m^*(\nu'))\right) \geq \frac{1}{4}\exp\left(-\sum_{a=1}^{K}
    \bE_\nu[N_a] \K(\nu_a,\nu'_a)\right).
  \]
\end{lemma}

\begin{figure}[b] \centering 
  \begin{center}
    \mbox{
      \includegraphics[height=3cm]{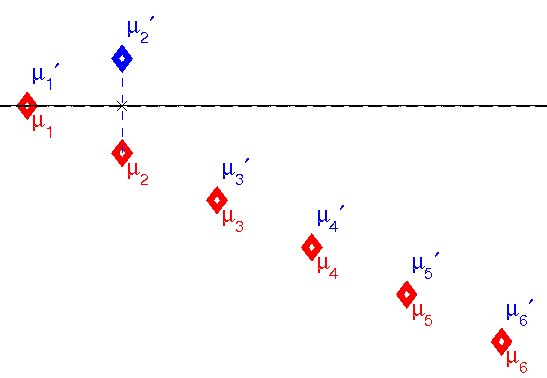}
      \hspace{1cm}
      \includegraphics[height=3cm]{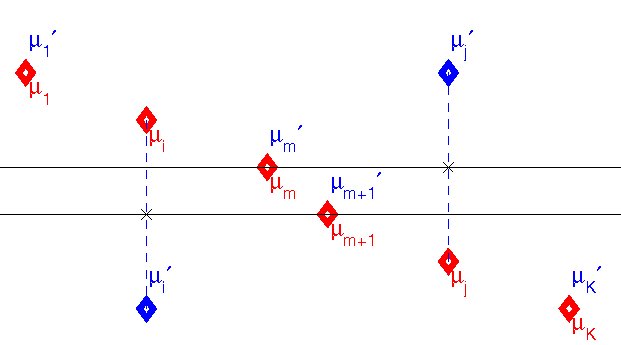}
    } \vspace{-0.4cm}
    \caption{Left: bandit models $\nu$, in red, and $\nu^{[2]}$, in blue (Theorem \ref{thm:FBGene1}).  Right: bandit models $\nu$, in red, and
      $\nu^{[i,j]}$, in blue (Theorem \ref{thm:FBGene2}).}
    \label{fig:Illustration}
    \vspace{-0.6cm}
  \end{center}
\end{figure}

\begin{theorem}\label{thm:FBGene1}
  Let $\nu$ be a Gaussian bandit model such that
  $\mu_1>\mu_2\geq\dots\geq\mu_K$ and let
  \[H'(\nu)=\sum_{a=2}^{K}\frac{2\sigma^2}{(\mu_1-\mu_a)^2}.\] 
  There exists a bandit model
  $\nu^{[a]}$, $a\in\{2,\dots,K\}$, (see Figure~\ref{fig:Illustration}) which satisfies $H'(\nu^{[a]})\leq H'(\nu)$ and is such
  that \[\max\left(p_t(\nu),p_t(\nu^{[a]})\right)\geq
  \exp\left(-\frac{4t}{H'(\nu)}\right).\]
\end{theorem}

This result is to be compared to the lower bound of \cite{Bubeck:BestArm10}. While Theorem~\ref{thm:FBGene1} does not really provide a lower bound on $\kappa_B(\nu)$, the complexity term $H(\nu)$ is close to the quantity that appears in Theorem~\ref{thm:GeneralBoundFC} for the fixed-confidence setting (in the Gaussian case), which improves over the term $H_2(\nu) = \max_{i: \mu_{[i]}<\mu_{[1]}} {i}{(\mu_{[1]}-\mu_{[i]})^{-2}}$ featured in Theorem 4 of \cite{Bubeck:BestArm10}.

For $m>1$, building on the same ideas, Theorem \ref{thm:FBGene2} provides a first lower
bound, which we believe leaves room for improvement.

\begin{theorem}\label{thm:FBGene2}
  Let $\nu$ be such that $\mu_1>\dots \mu_m>\mu_{m+1}>\dots>\mu_K$ and let
  \[H^+(\nu)=\sum_{a=1}^{m}\frac{2\sigma^2}{(\mu_a-\mu_{m+1})^2}, \ \ \
  H^-(\nu)=\sum_{a=m+1}^{K} \frac{2\sigma^2}{(\mu_{m}-\mu_a)^2}, \ \ \
  \text{and} \ \ \ H(\nu)=H^+(\nu)+H^-(\nu).\] There exists $a\in\{1,\dots,
  m\}$ and $b\in\{m+1,\dots K\}$ such that the bandit model $\nu^{[a,b]}$
  described on Figure \ref{fig:Illustration} satisfies $H(\nu^{[a,b]})<H(\nu)$
  and is such that
  \[\max\left(p_t(\nu),p_t(\nu^{[a,b]})\right) \geq
  \frac{1}{4}\exp\left(-\frac{4t}{\tilde{H}(\nu)}\right), \ \ \text{where} \ \
  \tilde{H}(\nu)=\frac{H(\nu) \min(H^+(\nu),H^-(\nu))}{H(\nu) +
    \min(H^+(\nu),H^-(\nu))}.\]
\end{theorem}

The proofs of Theorem~\ref{thm:FBGene1} and Theorem~\ref{thm:FBGene2} are very
similar. For this reason, we provide in Appendix \ref{proof:FBGene2} only the
latter. Introducing the gaps $\Delta_a$ defined in (\ref{def:HComplexity}), the
precise definition of the modified problems $\nu^{[a]}$ and $\nu^{[a,b]}$ in
the statement of the two results is:
\begin{equation*}
  \nu^{[a]}: \ 
  \left\{
    \begin{array}{ccl}
      \mu_k'&=&\mu_k \ \text{for all} \ k\neq a \\
      \mu_a'&=&\mu_a + 2\Delta_a 
    \end{array}
  \right.
  \ \ \text{and} \ \ \ \ 
  \nu^{[a,b]}: \ 
  \left\{
    \begin{array}{ccl}
      \mu_k'&=&\mu_k \ \text{for all} \ k\notin\{a,b\} \\
      \mu_a'&=&\mu_a - 2\Delta_b \\
      \mu'_b&=&\mu_b + 2\Delta_a
    \end{array}
  \right.
  .
\end{equation*}

\section{Numerical Experiments \label{sec:Experiments}}

In this section, we focus on two-armed models and provide experimental
experiments designed to compare the fixed-budget and fixed-confidence settings
(in the Gaussian and Bernoulli cases) and to illustrate the improvement
resulting from the adoption of the reduced exploration rate of Theorem
\ref{thm:PACFC}.

\begin{figure}[b] \centering \vspace{-0.3cm}
  \begin{center}
    \mbox{ \hspace{-0.2cm}
      \includegraphics[width=0.49\textwidth]{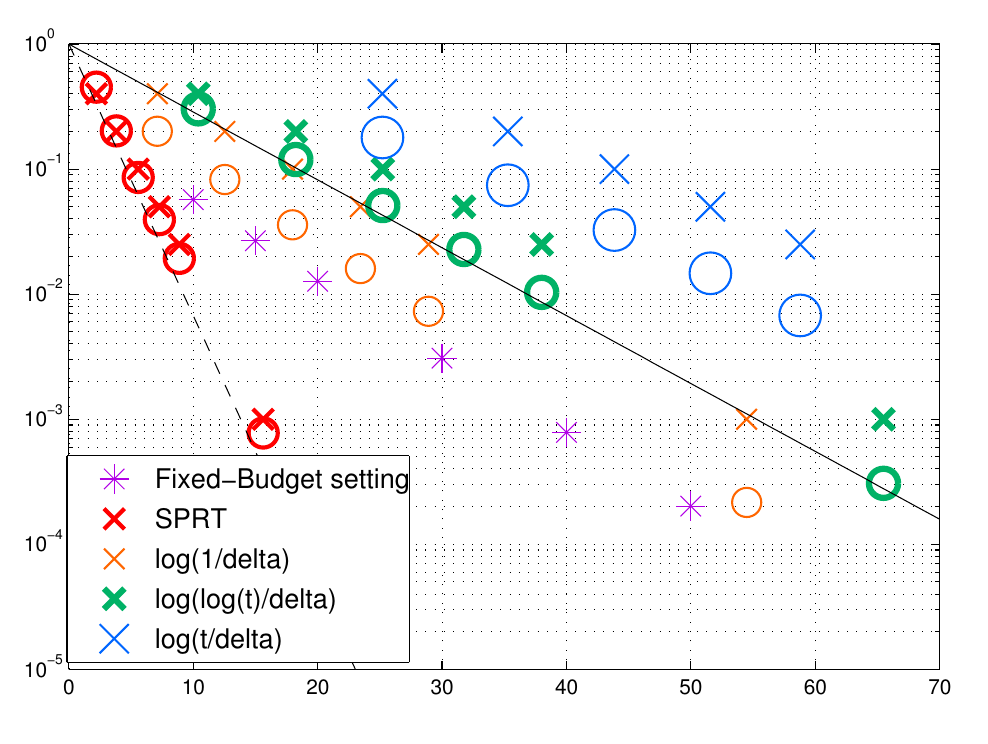}
      \hspace{-0.2cm}
      \includegraphics[width=0.49\textwidth]{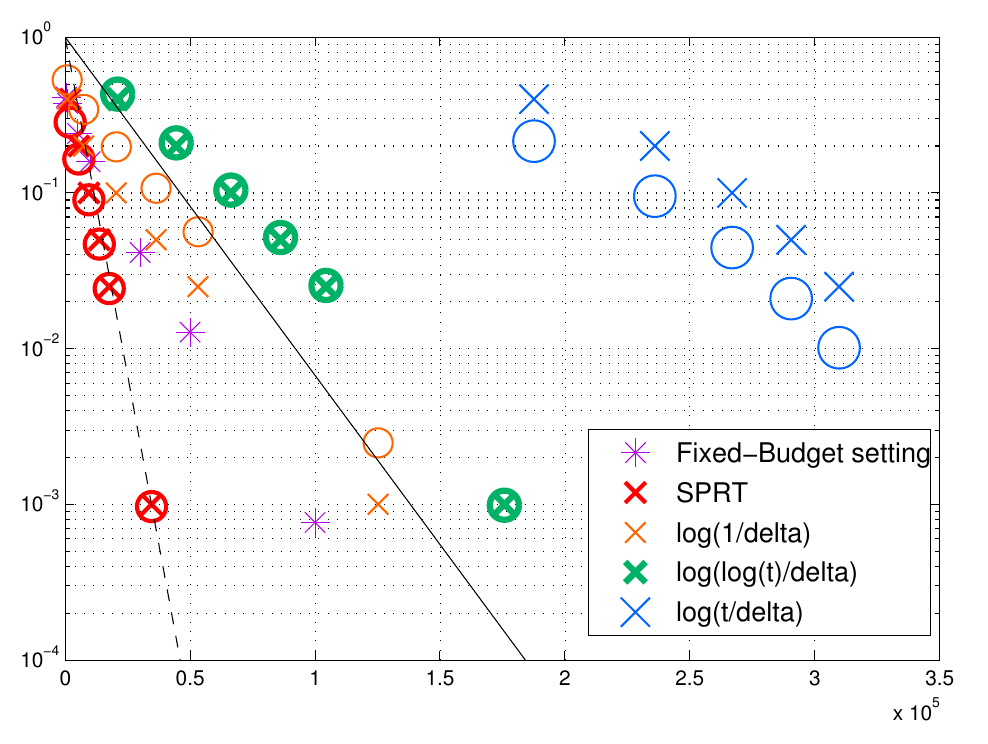}
    } \vspace{-0.4cm}
    \caption{Experimental results for Gaussian bandit
      models 
    }
    \label{fig:ResultsGaussian}
    \vspace{-0.6cm}
  \end{center}
\end{figure}

In Figure \ref{fig:ResultsGaussian}, we consider two Gaussian bandit models with known common variance:
the `easy' one is $\{\norm{0.5}{0.25}, \norm{0}{0.25}\}$, corresponding to $\kappa_C = \kappa_B = \kappa=8$, on the left; and the `difficult' one is $\{\norm{0.01}{0.25}, \norm{0}{0.25}\}$, that is
$\kappa=2 \times 10^4$, on the right. In the fixed-budget setting, stars ('*') report the
probability of error $p_n(\nu)$ as a function of $n$. In the fixed-confidence
setting, we plot both the empirical probability of error by circles ('O') and
the specified maximal error probability $\delta$ by crosses ('X') as a function
of the empirical average of the running times. Note the logarithmic scale used
for the probabilities on the y-axis.  All results are averaged over $N=10^6$
independent Monte Carlo replications. For comparison purposes, a plain line
represents the theoretical rate $t \mapsto \exp(-t (1/\kappa))$ which is a
straight line on the log scale.

In the fixed-confidence setting, we report results for elimination algorithms
of the form (\ref{Robbins}) for three different exploration rates
$\beta(t,\delta)$. The exploration rate we consider are: the provably-PAC rate
of Robbins' algorithm $\log({t}/{\delta})$ (large blue symbols), the
conjectured optimal exploration rate $\log({(\log(t)+1)}/{\delta})$, almost
provably $\delta$-PAC according to Theorem \ref{thm:PACFC} (bold green
symbols), and the rate $\log ({1}/{\delta})$, which would be appropriate if we
were to perform the stopping test only at a single pre-specified time (orange
symbols). For each algorithm, the log probability of error is approximately a
linear function of the number of samples, with a slope close to $-1/\kappa$,
where $\kappa$ is the complexity. {A first observation is that the 'traditional' 
rate of $\log({t}/{\delta})$ is much too conservative, with running times for
the difficult problem (right plot) which are about three times longer than
those of other methods for comparable error rates. As expected, the rate
$\log((\log(t)+1)/\delta)$ significantly reduces the running times while
maintaining proper control of the probability of failure, with empirical error
rates ('O' symbols) below the corresponding confidence parameters
$\delta$ (represented by 'X' symbols). Conversely, the use of the
non-sequential testing threshold $\log(1/\delta)$ seems too risky, as one can
observe that the empirical probability of error may be larger than $\delta$ on
difficult problems. To illustrate the gain in sample complexity resulting from
the knowledge of the means}, we also represented in red the performance of the
SPRT algorithm mentioned in the introduction of Section \ref{sec:FixedBudget}
along with the theoretical relation between the probability of error and the
expected number of samples, materialized as a dashed line. The SPRT stops for
$t$ such that $|(\mu_1-\mu_2)(S_{1,t/2} - S_{2,t/2})|>\log(1/\delta)$.

\begin{figure}[t]
  \vspace{-0.3cm}
  \begin{center}
    \mbox{ \hspace{-0.2cm}

      \includegraphics[width=0.49\textwidth]{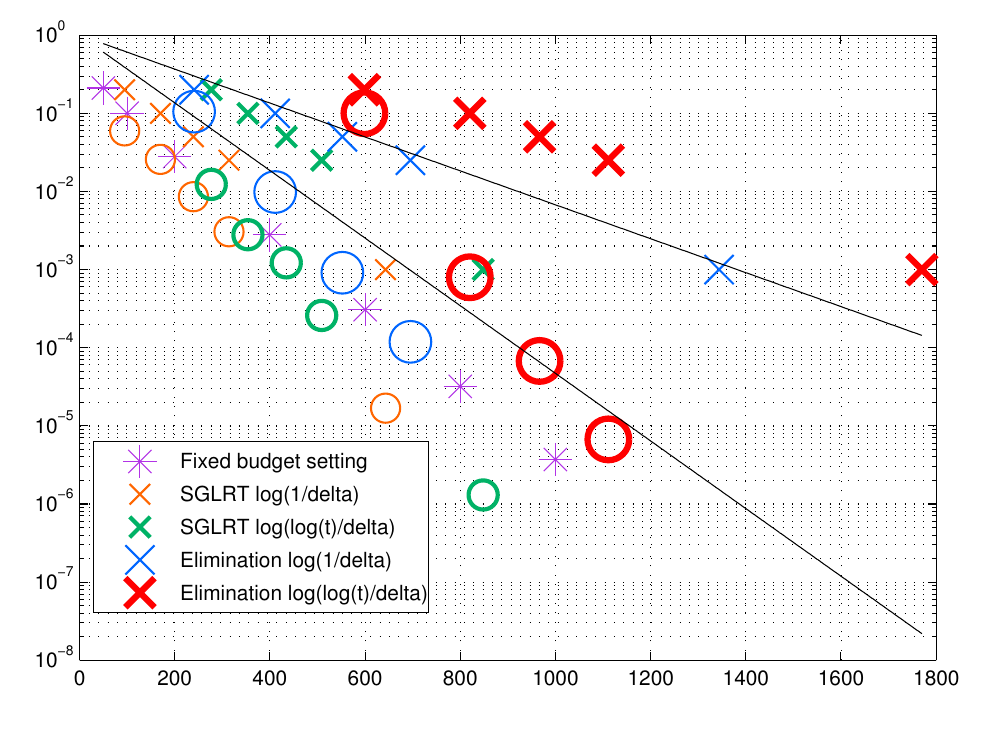}
      \hspace{-0.2cm}

\includegraphics[width=0.49\textwidth]{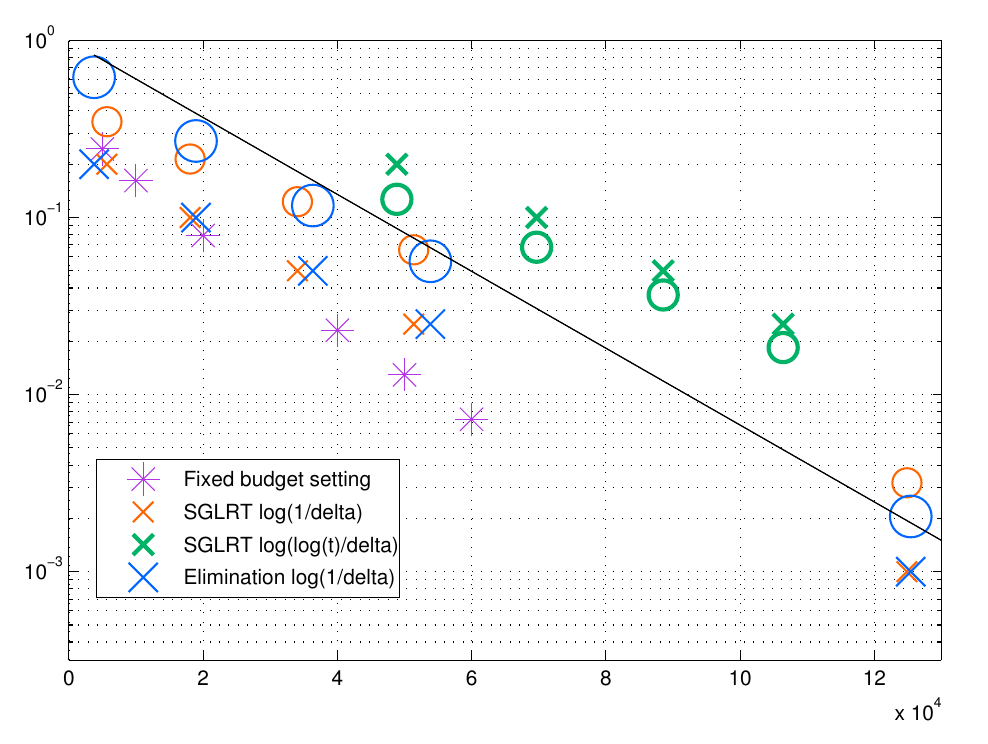}
}

\vspace{-0.4cm}
\caption{Results for Bernoulli bandit models: $0.2-0.1$ (left) and $0.51-0.5$
  (right).}
\label{fig:ResultsBernoulli}
\vspace{-0.6cm}
\end{center}
\end{figure}

Robbins' algorithm is $\delta$-PAC and matches the complexity (which is
illustrated by the slope of the measures), though in practice the use of the
exploration rate $\log((\log(t)+1)/\delta)$ leads to huge gain in terms of
number of samples used. It is important to keep in mind that running times play
the same role as error exponents and hence the threefold increase of average
running times observed on the rightmost plot of
Figure~\ref{fig:ResultsGaussian} when using
$\beta(t,\delta)=\log({t}/{\delta})$ is really prohibitive.

On Figure \ref{fig:ResultsBernoulli}, we compare on two Bernoulli bandit models
the performance of the SGLRT algorithm described in Section \ref{sec:Bernoulli} (Algorithm~\ref{AlgoBox:SGLRT}) using two different exploration rates, $\log(1/\delta)$ and
$\log((\log(t)+1)/\delta)$, to the 1/2-elimination stopping rule (Algorithm~\ref{AlgoBox:Elimination}) that stops when the
difference of empirical means exceeds the threshold $\sqrt{2\beta(t,\delta)/t}$
(for the same exploration rates). Plain lines also materialize the theoretical
optimal rate $t \mapsto \exp(-t/\kappa_C(\nu))$ and the rate attained by the
1/2-Elimination algorithm $t \mapsto \exp(-t/\kappa')$, where
$\kappa'=2/(\mu_1-\mu_2)^2$. On the bandit model $0.51-0.5$ (right) these two
rates are very close and SGLRT mostly coincides with Elimination, but on the
bandit model $0.2-0.1$ (left) the practical gain of the use of a more
sophisticated stopping strategy is well illustrated. Besides, our experiments
show that SGLRT using $\log((\log(t)+1)/\delta)$ is $\delta$-PAC on both the
(relatively) easy and difficult problems we consider, unlike the other
algorithms considered.

If one compares the results for the fixed-budget setting (in purple) to those
for the best $\delta$-PAC algorithm (or conjectured $\delta$-PAC for SGLRT in
the Bernoulli case), in green, one can observe that to obtain the same
probability of error, the fixed-confidence algorithm usually needs an average
number of samples that is about twice larger than the deterministic number of
samples required by the fixed-budget setting algorithm. This remark should be
related to the fact that a $\delta$-PAC algorithm is designed to be uniformly
good across all problems, whereas consistency is a weak requirement in the
fixed-budget setting: any strategy that draws both arm infinitely often and
recommends the empirical best is consistent. Figure~\ref{fig:ResultsGaussian}
also shows that when the values of $\mu_1$ and $\mu_2$ are unknown, the
sequential version of the test is no more preferable to its batch counterpart
and can even become much worse if the exploration rate $\beta(t,\delta)$ is
chosen too conservatively. This observation should be mitigated by the fact
that the sequential (or fixed-confidence) approach is adaptive with respect to
the difficulty of the problem whereas it is impossible to predict the
efficiency of a batch (or fixed-budget) experiment without some prior knowledge
regarding the difficulty of the problem under consideration.

\section{Conclusion}

 Our aim with this paper has been to provide a framework for evaluating, in a
principled way, the performance of fixed-confidence and fixed-budget algorithms
designed to identify the best arm(s) in stochastic environments.

For two-armed bandits, we obtained rather complete results,
identifying the complexity of both settings in
important parametric families of distributions. In doing so, we observed that
standard testing strategies based on uniform sampling are optimal or close to
optimal for Gaussian distributions with matched variance or Bernoulli
distributions but can be improved (by non-uniform sampling) for Gaussian
distributions with distinct variances. This latter observation can certainly be
generalized to other models, starting with the case of Gaussian distributions
whose variances are a priori unknown. In the case of Bernoulli distributions, we
have also shown that fixed-confidence algorithms that use the difference of the
empirical means as a stopping criterion are bound to be sub-optimal. Finally, we have
shown, through the comparison of the complexities $\kappa_C(\nu)$ and
$\kappa_B(\nu)$, that the behavior observed when testing fully specified
alternatives where fixed confidence (or sequential) algorithms may be `faster
on average' than the fixed budget (or batch) ones is not true anymore when the
parameters of the models are unknown.

For models with more than two arms, we obtained the first generic (i.e. not
based on the sub-Gaussian tail assumption) distribution-dependent lower bound
on the complexity of $m$ best-arms identification in the fixed-confidence
setting (Theorem~\ref{thm:GeneralBoundFC}). Currently available performance
bounds for algorithms performing $m$ best-arms identification---those of
\cite{COLT13} notably---show a small gap with this result and it is certainly
of interest to investigate whether those analyses and/or the bound of
Theorem~\ref{thm:GeneralBoundFC} may be improved to bridge the gap. For the
fixed-budget setting we made only a small step towards the understanding of the
complexity of $m$ best-arms identification and our results can certainly be greatly improved.

\acks{We thank S\'ebastien Bubeck for fruitful discussions during the visit of
  the first author at Princeton University. This work has been supported by the
  ANR-2010-COSI-002 and ANR-13-BS01-0005 grants of the French National Research Agency.}


\appendix

\section{Changes of Distributions \label{proofs:ChangeDist}}

Let $\nu$ and $\nu'$ be two bandit models such that for all $a \in \{1,K\}$ 
the distributions $\nu_a$ and $\nu_a'$ are mutually absolutely continuous. 
For each $a$, there exists a measure $\lambda_a$ such that $\nu_a$ and $\nu'_a$ 
have a density $f_a$ and $f'_a$ respectively with respect to $\lambda_a$. 
One can introduce the log-likelihood ratio of
the observations up to time $t$ under an algorithm $\cA$:
\[L_t = L_t(A_1,\dots,A_t,Z_1,\dots,Z_t) :=
\sum_{a=1}^{K}\sum_{s=1}^{t}\ind_{(A_s=a)}\log\left(\frac{f_{a}(Z_s)}{f'_a(Z_s)}\right).\]

The key element in a change of distribution is the following classical lemma
that relates the probabilities of an event under $\bP_\nu$ and $\bP_{\nu'}$
through the log-likelihood ratio of the observations. Such a result has often
been used in the bandit literature for $\nu$ and $\nu'$ that differ just from
one arm, for which the expression of the log-likelihood ratio is simpler. In
this paper, we consider more general changes of distributions, and we therefore
provide a full proof of Lemma \ref{Garivier} in Appendix~\ref{proof:Garivier}.
 
\begin{lemma}\label{Garivier} Let $\sigma$ be any stopping time with respect to
  $\mathcal{F}_t$.  For every event $\cE\in\mathcal{F}_\sigma$ (i.e., $\cE$ such
  that $\cE\cap (\sigma=t) \in \mathcal{F}_t$),
$$\bP_{\nu'}(\cE)=\bE_\nu[\ind_{\cE}\exp(-L_{\sigma})]$$
\end{lemma}

\subsection{Proof of Lemma \ref{lem:Cornerstone}}

To prove Lemma~\ref{lem:Cornerstone}, we state a first inequality on the expected log-likelihood ratio in Lemma~\ref{lem:CornerstoneLike}, which is of independent interest.

\begin{lemma}\label{lem:CornerstoneLike} Let $\sigma$ be any  almost surely finite stopping time with respect to
  $\mathcal{F}_t$.  For every event $\cE\in\mathcal{F}_\sigma$,
\[\bE_\nu[L_{\sigma}] \geq d(\bP_\nu(\cE),\bP_{\nu'}(\cE)).\] 
\end{lemma}

Lemma~\ref{lem:Cornerstone} easily follows: introducing $(Y_{a,s})$, the sequence of i.i.d. samples successively observed
from arm $a$, the log-likelihood ratio $L_t$ can be rewritten
\[L_t =
\sum_{a=1}^{K}\sum_{s=1}^{N_a(t)}\log\left(\frac{f_a(Y_{a,s})}{f'_a(Y_{a,s})}\right);
\ \ \ \text{and} \ \ \
\bE_\nu\left[\log\left(\frac{f_a(Y_{a,s})}{f'_a(Y_{a,s})}\right)\right]=\K(\nu_{a},\nu'_{a}).\]
Wald's Lemma (see e.g., \cite{Siegmund:SeqAn}) applied to 
$L_\sigma =\sum_{a=1}^{K}\sum_{s=1}^{N_a(\sigma)}\log\left(\frac{f_a(Y_{a,s})}{f'_a(Y_{a,s})}\right)$
yields
\begin{equation}\bE_\nu[L_\sigma] = \sum_{a=1}^{K}\bE_\nu[N_a(\sigma)]\K(\nu_{a},\nu'_{a}).\label{equ:Wald}\end{equation}
Combining this equality with the inequality in Lemma~\ref{lem:CornerstoneLike} completes the proof. 

\medskip

\emph{Proof of Lemma~\ref{lem:CornerstoneLike}.}
Let $\sigma$ be a stopping time with respect to $(\cF_t)$.  

We start by showing
that for all $\cE\in \cF_\sigma$, $\bP_\nu(\cE)=0 \Leftrightarrow \bP_{\nu'}(\cE)=0$. This proves Lemma~\ref{lem:CornerstoneLike} for events $\cE$ such that $\bP_\nu(\cE)=0$ or $1$, 
for which the quantity $d(\bP_\nu(\cE),\bP_{\nu'}(\cE)) = d(0,0)$ or $d(1,1)$ is equal to zero by convention, and the inequality thus holds since the left-hand side is non-negative (which is clear from the 
rewriting \eqref{equ:Wald}). Let $\cE\in \cF_\sigma$. Lemma \ref{Garivier} yields
$\bP_{\nu'}(\cE) = \bE_\nu[\ind_{\cE}\exp(-L_\sigma)]$.  Thus $\bP_{\nu'}(\cE)=0$
implies $\ind_{\cE}\exp(-L_\sigma)=0 \ \bP_\nu-a.s$.  As $\bP_\nu(\sigma < +
\infty)=1$, $\bP_\nu(\exp(L_\sigma) > 0)=1$ and $\bP_{\nu'}(\cE)=0 \Rightarrow
\bP_\nu(\cE)=0$. A similar reasoning yields $\bP_\nu(\cE)=0 \Rightarrow
\bP_{\nu'}(\cE)=0$.

Let $\cE\in\cF_\sigma$ be such that $0<\bP_\nu(\cE)<1$ (then
$0<\bP_{\nu'}(\cE)<1$). Lemma \ref{Garivier} and the conditional Jensen
inequality lead to
\begin{align*}
  \bP_{\nu'}(\cE) & = \bE_\nu[\exp(-L_\sigma)\ind_{\cE}] =  \bE_\nu[\bE_\nu[\exp(-L_\sigma) | \ind_{\cE}] \ind_{\cE}] \\
  & \geq \bE_\nu[\exp\left(-\bE_\nu[L_\sigma | \ind_{\cE}]\right) \ind_{\cE}]  = \bE_\nu[\exp\left(-\bE_\nu[L_\sigma | \ind_{\cE}]\ind_{\cE}\right) \ind_{\cE}]\\
  & =  \bE_\nu[\exp\left(-\bE_\nu[L_\sigma |\cE]\ind_{\cE}\right) \ind_{\cE}] =  \bE_\nu[\exp\left(-\bE_\nu[L_\sigma |\cE]\right) \ind_{\cE}]\\
  & = \exp\left(-\bE_\nu[L_\sigma | \cE]\right) \bP_\nu(\cE).
\end{align*}
Writing the same for the event $\overline{\cE}$ yields
$\bP_{\nu'}(\overline{\cE})\geq \exp\left(-\bE_\nu[L_\sigma |
  \overline{\cE}]\right) \bP_\nu(\overline{\cE})$, hence
\begin{equation}
  \bE_\nu[L_\sigma | \cE] \geq \log \frac{\bP_\nu(\cE)}{\bP_{\nu'}(\cE)} \ \ \ \text{and} \ \ \ \bE_\nu[L_\sigma | \overline{\cE}] \geq \log \frac{\bP_\nu(\overline{\cE})}{\bP_{\nu'}(\overline{\cE})}. 
  \label{Similar}
\end{equation}
Therefore one can write
\begin{eqnarray}
  \bE_\nu[L_\sigma ] & = & \bE_\nu[L_\sigma | \cE]\bP_\nu(\cE) + \bE_\nu[L_\sigma | \overline{\cE}]\bP_\nu(\overline{\cE}) \nonumber \\
  & \geq & \bP_\nu(\cE) \log \frac{\bP_\nu(\cE)}{\bP_{\nu'}(\cE)} + \bP_\nu(\overline{\cE})\log \frac{\bP_\nu(\overline{\cE})}{\bP_{\nu'}(\overline{\cE})} = d(\bP_\nu(\cE),\bP_{\nu'}(\cE)) \nonumber,
\end{eqnarray}
which concludes the proof. 

\subsection{Proof of Lemma \ref{lem:CornerVariant}\label{proof:CornerVariant}}

The proof bears strong similarities with that of Lemma \ref{lem:Cornerstone},
but an extra ingredient is needed: Lemma 4 of \cite{Bubeck:Rigollet13}, that
provides a lower bound on the sum of type I and type II probabilities of error
in a statistical test.
\begin{lemma} \label{Rigollet}Let $\rho_0$,$\rho_1$ be two probability
  distributions supported on some set $\mathcal{X}$, with $\rho_1$ absolutely
  continuous with respect to $\rho_0$. Then for any measurable function $\phi :
  \mathcal{X} \rightarrow \{0,1\}$, one has
  \[\bP_{X\sim \rho_0}(\phi(X)=1) + \bP_{X\sim \rho_1}(\phi(X)=0) \geq
  \frac{1}{2}\exp(-\K(\rho_0,\rho_1)).\]
\end{lemma}
Let $\nu$ and $\nu'$ be two bandit models that do not have the same set of
optimal arms.  We denote by $\cS_1,\dots, \cS_M$ the $M={K \choose m}$ subsets
of $m$, ordered so that $\cS_1$ (resp. $\cS_2$) is the set of $m$ best arms in
problem $\nu$ (resp. $\nu'$). One has
\begin{eqnarray*}
  \max\left(\bP_\nu(\hat{\cS}_m\neq \cS_1),\bP_{\nu'}(\hat{\cS}_m\neq \cS_2)\right) 
  & \geq & \frac{1}{2}\left(\bP_\nu(\hat{\cS}_m\neq \cS_1) + \bP_{\nu'}(\hat{\cS}_m\neq \cS_2)\right) \\
  &\geq &\frac{1}{2}\left(\bP_\nu(\hat{\cS}_m\neq \cS_1) + \bP_{\nu'}(\hat{\cS}_m = \cS_1)\right).
\end{eqnarray*}
Let $\rho_0=\mathcal{L}(\hat{\cS}_m)$ and $\rho_1=\mathcal{L}'(\hat{\cS}_m)$ be
the distribution of $\hat{\cS}_m$ for algorithm $\mathcal{A}$ under problems
$\nu$ and $\nu'$ respectively.  $\rho_1$ is absolutely continuous with respect
to $\rho_0$, since as mentioned above, for any event in $\cF_t$, $\bP_\nu(A)=0
\Leftrightarrow \bP_{\nu'}(A)=0$.  Therefore one can apply Lemma \ref{Rigollet}
with $\rho_0$,$\rho_1$ and $\phi(x)=\ind_{(x\neq \cS_1)}$ and write
\[\max\left(\bP_\nu(\hat{\cS}_m\neq \cS_1),\bP_{\nu'}(\hat{\cS}_m\neq
  \cS_2)\right) \geq
\frac{1}{4}\exp\left(-\K(\mathcal{L}(\hat{\cS}_m),\mathcal{L}'(\hat{\cS}_m))\right).\]
To conclude the proof, it remains to show that
$\K(\mathcal{L}(\hat{\cS}_m),\mathcal{L}'(\hat{\cS}_m))$ is upper bounded by
$\sum_{a=1}^{K} \bE_\nu[N_a(t)] \K(\nu_{a},\nu'_{a})$, which is equal to
$\bE_\nu[L_t]$, as shown above (equation (\ref{equ:Wald})).

The rest of the proof boils down to prove a lower bound on $\bE_\nu[L_t]$
slightly different from the one used to obtain Lemma \ref{lem:Cornerstone}.
For $k\in\{1,\dots,M\}$, applying inequality (\ref{Similar}) to
$(\hat{\cS}_m=\cS_k)\in\cF_\tau$ yields
\begin{equation*} \bE_\nu[L_t | \hat{\cS}_m=\cS_k] \geq
  \log\left(\frac{\bP_\nu(\hat{\cS}_m=\cS_k)}{\bP_{\nu'}(\hat{\cS}_m=\cS_k)}\right).\end{equation*}
Thus one can write, letting $\mathcal{I}=\{k\in\{1,\dots,M\} :
\bP_\nu(\hat{\cS}_m=\cS_k)\neq 0\}$,
\begin{eqnarray*}
  \bE_\nu[L_t] & = & \sum_{k\in\mathcal{I}} \bE_\nu[L_t| \hat{\cS}_m=\cS_k]\bP(\hat{\cS}_m=\cS_k) \\
  & \geq  &\sum_{k\in\mathcal{I}}\log\left(\frac{\bP_\nu(\hat{\cS}_m=\cS_k)}{\bP_{\nu'}(\hat{\cS}_m=\cS_k)}\right)\bP_\nu(\hat{\cS}_m=\cS_k) 
  = \K(\mathcal{L}(\hat{\cS}_m),\mathcal{L}'(\hat{\cS}_m)),
\end{eqnarray*}
which concludes the proof.

\subsection{Proof of Lemma \ref{Garivier} \label{proof:Garivier}}
Recall that for all $a \in \{1,\dots,K\}$ there exists a measure $\lambda_a$ such that $\nu_a$ (resp. $\nu_a'$) has density
$f_a$ (resp. $f'_a$) with respect to $\lambda_a$.  For all $a\in\{1,\dots,K\}$, let
$(Y_{a,t})_{t\in\N}$ be an i.i.d. sequence such that if $A_t=a$, $Z_t=Y_{a,t}$.

We start by showing by induction that for all $n\in \N$ the following statement
is true: for every function $g:\R^n \rightarrow \R$ measurable,
\[\bE_{\nu'}[g(Z_1,\dots,Z_n)] =
\bE_\nu\left[g(Z_1,\dots,Z_n)\exp(-L_n(Z_1,\dots,Z_n))\right].\] The result for
$n=1$ follows from the following calculation:
\begin{eqnarray*}
  \bE_{\nu'}[g(Z_1)] & = & \bE_{\nu'}\left[\sum_{a=1}^K\ind_{(A_1=a)}g(Y_{a,1})\right] = \sum_{a=1}^K\bE_{\nu'}\left[\ind_{(A_1=a)}\bE_{\nu'}[g(Y_{a,1})|\cF_0]\right]  \\ 
  & = & \sum_{a=1}^K \bP_{\nu'}(A_1=a)\bE_{\nu'}[g(Y_{a,1})]  = \sum_{a=1}^K \bP_\nu(A_1=a)\bE_\nu\left[g(Y_{a,1})\frac{f'_a(Y_{a,1})}{f_a(Y_{a,1})}\right] \\
  & = & \bE_\nu \left[\sum_{a=1}^K \ind_{(A_1=a)} g(Y_{a,1}) \frac{f'_a(Y_{a,1})}{f_a(Y_{a,1})}\right] \\ 
  &=& \bE_\nu \left[g(Z_1)\sum_{a=1}^K \ind_{(A_1=a)} 
    \exp\left(- \log\frac{f'_a(Z_1)}{f_a(Z_1)}\right)\right]\\
  & = & \bE_\nu \left[g(Z_1)\exp\left(- \sum_{a=1}^K\ind_{(A_1=a)}\log\frac{f'_a(Z_1)}{f_a(Z_1)}\right)\right] \\ &=& \bE_\nu\left[g(Z_1)\exp(-L_1(Z_1))\right].
\end{eqnarray*}
We use that the initial choice of action satisfies
$\bP_\nu(A_1=a)=\bP_{\nu'}(A_1=a)$.

We now assume that the statement holds for some integer $n$, and show it holds
for $n+1$. Let $g:\R^{n+1} \rightarrow \R$ be a measurable function.
\begin{align*}
  & \bE_{\nu'}[g(Z_1,\dots,Z_n,Z_{n+1})] = \bE_{\nu'}[\bE_{\nu'}[g(Z_1,\dots,Z_n,Z_{n+1}) | \cF_n]]  \\
  &\hspace{1.5cm} \overset{(*)}{=}   \bE_\nu \left[\bE_{\nu'}[g(Z_1,\dots,Z_n,Z_{n+1}) | \cF_n]\exp\left(-L_n(Z_1,\dots,Z_n)\right)\right] \\
  &\hspace{1.5cm} =  \bE_\nu\left[\sum_{a=1}^{K} \ind_{A_{n+1}=a} \bE_{\nu'}[g(Z_1,\dots,Z_n,Y_{a,n+1}) | \cF_n]\exp\left(-L_n(Z_1,\dots,Z_n)\right)\right] \\
  &\hspace{1.5cm} = \bE_\nu\left[\sum_{a=1}^{K} \ind_{A_{n+1}=a} \int
    g(Z_1,\dots,Z_n,z)
    \frac{f'_a(z)}{f_a(z)}f_a(z)d\lambda_a(z)\exp\left(-L_n(Z_1,\dots,Z_n)\right)\right].
\end{align*}
Observing that on the event $(A_{n+1}=a)$, $L_{n+1}(Z_1,\dots,Z_n,z)=
L_n(Z_1,\dots,Z_n) + \log \frac{f_a(z)}{f'_a(z)}$ leads to:
\begin{align*}
  & \bE_{\nu'}[g(Z_1,\dots,Z_n,Z_{n+1})] \\
  &\hspace{1.5cm}=  \bE_\nu\left[\sum_{a=1}^{K} \ind_{A_{n+1}=a} \int g(Z_1,\dots,Z_n,z) \exp(-L_{n+1}(Z_1,\dots,Z_n,z))f_a(z)d\lambda_a(z)\right]\\
  &\hspace{1.5cm} = \bE_\nu\left[\sum_{a=1}^{K} \ind_{A_{n+1}=a}\bE_\nu\left[ g(Z_1,\dots,Z_n,Y_{a,n+1}) \exp(-L_{n+1}(Z_1,\dots,Z_n,Y_{a,n+1}))|\cF_{n}\right]\right] \\
  &\hspace{1.5cm} =  \bE_\nu\left[\bE_\nu\left[g(Z_1,\dots,Z_n,Z_{n+1})\exp(-L_{n+1}(Z_1,\dots,Z_n,Z_{n+1}))|\cF_{n}\right]\right] \\
  &\hspace{1.5cm} =
  \bE_\nu\left[g(Z_1,\dots,Z_n,Z_{n+1})\exp(-L_{n+1}(Z_1,\dots,Z_n,Z_{n+1})\right].
\end{align*}
Hence, the statement is true for all $n$, and we have shown that for every $\cE
\in \cF_n$,
$$\bP_{\nu'}(\cE)=\bE_\nu[\ind_{\cE}\exp(-L_n)].$$
Let $\sigma$ be a stopping time w.r.t. $ (\cF_n)$ and $\cE\in \cF_\sigma$.
\begin{eqnarray*}
  \bP_{\nu'}(\cE) & = & \bE_{\nu'}[\ind_{\cE}] = \sum_{n=0}^{\infty}\bE_{\nu'}[\underbrace{\ind_{\cE} \ind_{(\sigma = n)}}_{\in \cF_n}]  = \sum_{n=0}^\infty \bE_\nu[\ind_{\cE}\ind_{(\sigma=n)}\exp(-L_n)] = \bE_\nu[\ind_{\cE}\exp(-L_\sigma)].
\end{eqnarray*}

\section{A Short Proof of Burnetas and Katehakis' Lower Bound on the
  Regret\label{proof:LaiRobbins}}

In the regret minimization framework, briefly described in the Introduction, a
bandit algorithm only consists in a sampling rule (there is no stopping rule
nor recommendation rule). The arms must be chosen sequentially so as to
minimize the regret, that is strongly related to the number of draws of the
sub-optimal arms (using the notation $\mu^* = \mu_{[1]}$):
\begin{eqnarray}
  R_T(\nu) & = & \mu^*T - \bE_\nu\,\left[\sum_{t=1}^T Z_t\right]  =  \sum_{a : \mu_a < \mu^*} (\mu^* - \mu_a)\,\bE_\nu\big[N_a(T)\big] \label{RegretNumber}
\end{eqnarray}
The lower bound given by \cite{LaiRobbins85bandits} on the regret holds for
families of distributions parameterized by a (single) real parameter. Their
result has been generalized by \cite{Burn:Kat96} to larger classes of
parametric distributions.  The version we give here deals with identifiable
classes of the form $\cM=(\cP)^K$, where $\cP$ is a set of probability measures
satisfying
\[
\forall \nu_a,\nu_b\in\cP, \ \nu_a\neq \nu_b \ \Rightarrow \ 0< \K(\nu_a,\nu_b)
< + \infty.
\]
\begin{theorem}\label{thm:LaiRobbinsPlus}
  Let $\cM$ be an identifiable class of bandit models. Consider a bandit
  algorithm such that for all $\nu\in\cM$ having a unique optimal arm, for all
  $\alpha\in (0,1]$, $R_T(\nu) = o(T^\alpha)$.  Then, for all $\nu\in\cM$,
  \begin{equation}\label{ineq:LR2}
    \mu_a < \mu^* \ \Rightarrow \ \liminf_{T\rightarrow \infty}\frac{\bE_\nu[N_a(T)]}{\log(T)}\geq \frac{1}{\Kinf\,(\nu_a;\mu^*)},
  \end{equation}
  where $\Kinf\,(p;\mu)=\inf\left\{\K(p,q) : q\in\cP \ \text{and} \ \bE_{X\sim
      q}[X]>\mu\right\}.$
\end{theorem}

\emph{Proof.} Let $\nu=(\nu_1,\dots,\nu_K)$ be a bandit model such that arm $1$ is the unique
optimal arm. Without loss of generality, we show that inequality
(\ref{ineq:LR2}) holds for the sub-optimal arm $a=2$.  Consider the alternative
bandit model $\nu'$ such that $\nu'_a=\nu_a$ for all $a\neq 2$ and
$\nu'_2\in\cP$ is such that $\bE_{X\sim \nu'_2}[X]>\mu_1$. Arm 1 is thus the
unique optimal arm under the model $\nu$, whereas arm 2 is the unique optimal
arm under the model $\nu'$. For every integer $T$, let $\cE_T$ be the event
defined by
\[\cE_{T} =\left(N_1(T) \leq T - \sqrt{T}\right).\] Clearly, $\cE_T\in\cF_T$. From
Lemma \ref{lem:Cornerstone}, applied to the stopping time $\sigma=T$ a.s.,
\begin{equation}
  \bE_\nu[N_2(T)]\K(\nu_2,\nu'_2) \geq d(\bP_\nu(\cE_{T}),\bP_{\nu'}(\cE_{T})).\label{ineq:LR}
\end{equation}
The event $\cE_T$ is not very likely to hold under the model $\nu$, in which the
optimal arm should be drawn of order $T-C\log(T)$ times, whereas it is very
likely to happen under $\nu'$, in which arm 1 is sub-optimal and thus only
drawn little. More precisely, Markov inequality yields
\begin{eqnarray*}
  \bP_\nu(\cE_{T}) & = & \bP_\nu(T-N_1(T) \geq \sqrt{T}) \leq \frac{\sum_{a\neq 1}\bE_\nu[N_a(T)]}{\sqrt{T}} \\
  \bP_{\nu'}(\cE_{T}^c) & = & \bP_{\nu'}(N_1(T) \geq T - \sqrt{T}) \leq \frac{\bE_{\nu'}[N_1(T)]}{T-\sqrt{T}} \leq  \frac{\sum_{a\neq 2}\bE_{\nu'}[N_a(T)]}{T-\sqrt{T}} 
\end{eqnarray*}
From the formulation (\ref{RegretNumber}), every algorithm that is uniformly
efficient in the above sense satisfies
\[\sum_{a\neq 1}\bE_\nu[N_a(T)]=o(T^\alpha) \ \ \text{and} \ \ \sum_{a\neq
  2}\bE_{\nu'}[N_a(T)]=o(T^\alpha)\] for all $\alpha\in (0,1]$. Hence
$\bP_\nu(\cE_{T}) \underset{n\rightarrow \infty}{\rightarrow} 0 \ \ \ \text{and}
\ \ \ \bP_{\nu'}(\cE_{T}) \underset{n\rightarrow \infty}{\rightarrow} 1.$
Therefore, we get
\begin{eqnarray*}
  \frac{d(\bP_\nu(\cE_{T}),\bP_{\nu'}(\cE_{T}))}{\log(T)} \underset{T\rightarrow \infty}{\sim} \frac{1}{\log(T)}\log\left(\frac{1}{\bP_{\nu'}(\cE_{T}^c)}\right)
  \geq \frac{1}{\log(T)}\log\left(\frac{T-\sqrt{T}}{\sum_{a\neq 2}\bE_{\nu'}[N_a(T)]}\right).
\end{eqnarray*}
The right hand side rewrites
\[1 + \frac{\log\left(1-\frac{1}{\sqrt{T}}\right)}{\log(T)} -
\frac{\log\left(\sum_{a\neq 2}\bE_{\nu'}[N_a(T)]\right)}{\log(T)} \Tgoesto 1\]
using the fact that $\sum_{a\neq 2}\bE_{\nu'}[N_a(T)]=o(T^\alpha)$ for all
$\alpha\in (0,1]$.  Finally, for every $\nu'_2\in\cP$ such that
$\bE_{X\sim\nu'_2}[X]>\mu_1$ on obtains, using inequality (\ref{ineq:LR})
\[\liminf_{T\rightarrow \infty} \frac{\bE[N_2(T)]}{\log(T)} \geq
\frac{1}{\K(\nu_2,\nu'_2)}.\] For all $\epsilon\in(0,1)$, $\nu'_2$ can then be chosen
such that $\K(\nu_2,\nu'_2)\leq \Kinf(\nu_2,\mu_1)/(1-\epsilon)$, and the
conclusion follows when $\epsilon$ goes to zero.

\section{Properties of $\Kb_*$ and $\Kb^*$ in Exponential Families \label{sec:prop_k_star}}
In this section, we review properties of $\Kb_*$ defined in
Section~\ref{sec:2arms} as well as those of $\Kb^*$ defined
in Section~\ref{sec:FixedBudget} in the case of one-parameter exponential family
distributions. We recall that $\Kb_*(\theta_1,\theta_2)=\Kb(\theta_1,\theta_*)$ where $\theta_*$ is defined by 
$\Kb(\theta_1,\theta_*)=\Kb(\theta_2,\theta_*)$
and that $\Kb^*(\theta_1,\theta_2)=\Kb(\theta^*,\theta_1)$ where $\theta^*$ is defined by 
$\Kb(\theta^*,\theta_1)=\Kb(\theta^*,\theta_2).$

\begin{figure}[h]
  \centering
  \includegraphics[width=0.7\textwidth]{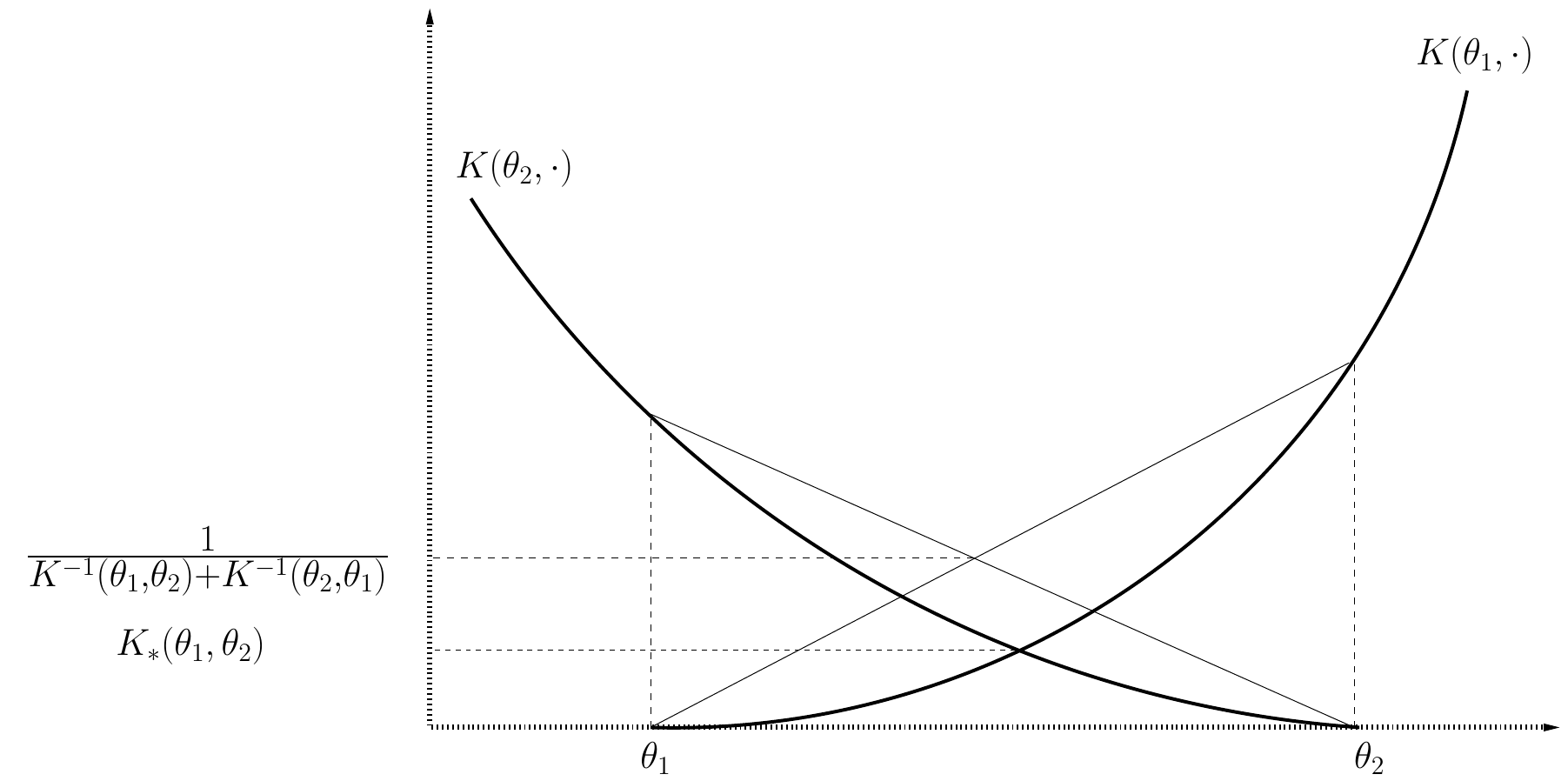}
  \caption{Comparison of the complexity terms featured in Theorems~\ref{thm:GeneralBoundFC} and~\ref{thm:2armsFC}.}
  \label{fig:k_star}
\end{figure}

Figure~\ref{fig:k_star} displays the geometric constructions corresponding to the complexity terms of Theorems~\ref{thm:GeneralBoundFC} and~\ref{thm:2armsFC}, respectively. As seen on the picture, the convexity of the function $\theta \mapsto K(\theta_i,\theta)$, for any value of $\theta_i$, {implies that 
\[\frac{1}{\Kb_*(\theta_1,\theta_2)} \geq \frac{1}{\Kb(\theta_1,\theta_2)} + \frac{1}{\Kb(\theta_2,\theta_1)}.\]}

\begin{figure}[h]
  \centering
  \includegraphics[width=0.35\textwidth]{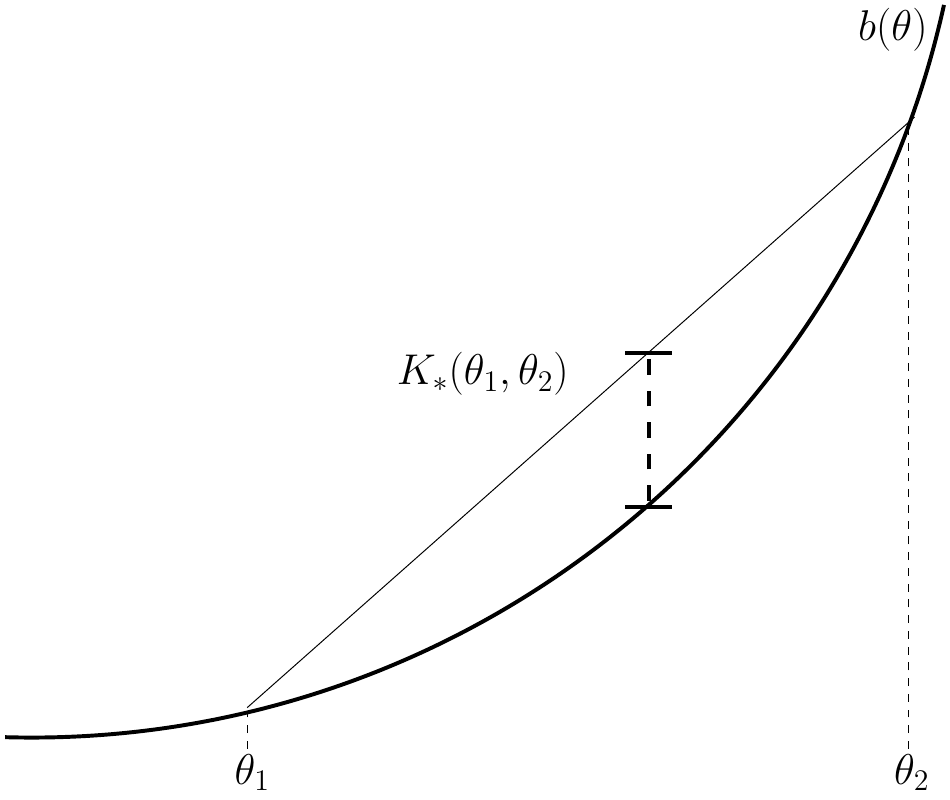}    
  \caption{Interpretation of $\Kb^*(\theta_1,\theta_2)$.}
  \label{fig:k_star_bregman}
\end{figure}

It is well known that in exponential families, the Kullback-Leibler divergence between distributions parameterized by their natural parameter, $\theta$, may be related to the Bregman divergence associated with the log-partition function $b$:
\[
\Kb(\theta_1,\theta_2) = b(\theta_2)-b(\theta_1) -\dot{b}(\theta_1)(\theta_2-\theta_1) = \operatorname{Bregman}_b(\theta_2,\theta_1) .
\]
From this representation, its is straightforward to show that
\begin{itemize}
\item $\Kb(\theta_1,\theta_2)$ is a twice differentiable strictly convex function of its second argument,
\item $\theta^*$ corresponds to the dual parameter $\mu^* := \dot{b}(\theta^*) = (b(\theta_2)-b(\theta_1))/(\theta_2-\theta_1)$,
\item $\Kb^*(\theta_1,\theta_2)$ admits the following variational representation
\[
  \Kb^*(\theta_1,\theta_2) = \max_{\theta\in(\theta_1,\theta_2)}\left\{b(\theta_1) + \frac{b(\theta_2)-b(\theta_1)}{\theta_2-\theta_1}(\theta-\theta_1) - b(\theta), \right\} ,
\]
corresponding to the maximal gap shown on Figure~\ref{fig:k_star_bregman} (achieved in $\theta^*$ for which $\dot{b}(\theta^*) = \mu^*$). The quantity $I^*(\theta_1,\theta_2)$ related to the use of uniform sampling, is equal to the value of the gap in $\theta = (\theta_1+\theta_2)/2$, which confirms that it is indeed smaller than $\Kb^*(\theta_1,\theta_2)$.
\end{itemize}

Indexing the distributions in the exponential family by their mean $\mu = \dot{b}(\theta)$ rather than their natural parameter $\theta$ and using the dual representation
\[
\Kb(\mu_1,\mu_2) = b^\star(\mu_1)-b^\star(\mu_2) -\dot{b}^\star(\mu_2)(\mu_1-\mu_2) = \operatorname{Bregman}_{b^\star}(\mu_1,\mu_2),
\]
where $b^\star(\mu):=\sup_{\theta} (\theta \mu - b(\theta))$ is the Fenchel conjugate of $b$, similarly yields
\begin{itemize}
\item $\Kb(\mu_1,\mu_2)$ is a twice differentiable strictly convex function of its first argument,
\item $\theta_* = \dot{b}^\star(\mu_*) = (b^\star(\mu_2)-b^\star(\mu_1))/(\mu_2-\mu_1)$;
\item $\Kb_*(\theta_1,\theta_2)$ is defined by
\[
  \Kb_*(\theta_1,\theta_2) = \max_{\mu\in(\mu_1,\mu_2)}\left\{b^\star(\mu_1) + \frac{b^\star(\mu_2)-b^\star(\mu_1)}{\mu_2-\mu_1}(\mu-\mu_1) - b^\star(\mu), \right\} .
\]
\end{itemize}

From what precedes, equality between $\Kb_*$ and $\Kb^*$ for all values of the parameters is only achievable when the log-partition function $b$ is self-conjugate.

\section{Proof of Theorem \ref{thm:MatchingFC} \label{proof:MatchingFC}}

Let $\alpha = {\sigma_1}/{(\sigma_1+\sigma_2)}$. We first prove that with the
exploration rate $\beta(t,\delta)=\log(t/\delta) + 2\log\log(6t)$ the algorithm
is $\delta$-PAC.  Assume that $\mu_1>\mu_2$ and recall $\tau=\inf\{t\in\N :
|d_t| > \sqrt{2\sigma^2_t(\alpha)\beta(t,\delta)}\}$, where
$d_t:=\hat{\mu}_1(t) - \hat{\mu}_2(t)$.  The probability of error of the
$\alpha$-elimination strategy is upper bounded by
\begin{eqnarray*}
  \bP_\nu\left(d_\tau \leq -\sqrt{{2\sigma_\tau^2(\alpha) \beta(\tau,\delta)}} \right) &\leq& \bP_\nu\left(d_\tau-(\mu_1-\mu_2) \leq - \sqrt{{2\sigma_\tau^2(\alpha) \beta(\tau,\delta)}} \right) \\
  &\leq& \bP_\nu\left(\exists t\in \N^* : d_t - (\mu_1 - \mu_2) <  - \sqrt{{2\sigma_t^2(\alpha) \beta(t,\delta)}}\right)\\
  & \leq & \sum_{t=1}^{\infty} \exp\left(-\beta(t,\delta)\right),
\end{eqnarray*}
by an union bound and Chernoff bound applied to $d_t - (\mu_1 - \mu_2)\sim
\norm{0}{\sigma^2_t(\alpha)}$.  The choice of $\beta(t,\delta)$ mentioned above
ensures that the series in the right hand side is upper bounded by $\delta$,
which shows the algorithm is $\delta$-PAC:
\begin{eqnarray*}\sum_{t=1}^{\infty} e^{-\beta(t,\delta)}&\leq& \delta \sum_{t=1}^{\infty}
\frac{1}{t(\log(6t))^2}\leq {\delta}\left(\frac{1}{(\log 6)^2} +
  \int_{1}^{\infty}\frac{dt}{t(\log(6t))^2}\right) \\
&=&{\delta}\left(\frac{1}{(\log 6)^2} + \frac{1}{\log(6)}\right)\leq \delta.
\end{eqnarray*}

To upper bound the expected sample complexity, we start by upper bounding the
probability that $\tau$ exceeds some deterministic time $T$:
\begin{eqnarray*}
  \bP_\nu(\tau \geq T) &\leq& \bP_\nu\left(\forall t =1\dots T, \ d_t \leq \sqrt{{2\sigma_t^2(\alpha)\beta(t,\delta)}}\right) \leq \bP_\nu\left(d_T \leq \sqrt{{2\sigma_T^2(\alpha)\beta(T,\delta)}}\right) \\
  &=& \bP_\nu\left( d_T - (\mu_1 - \mu_2) \leq -\left[(\mu_1 - \mu_2)-\sqrt{{2\sigma_T^2(\alpha)\beta(T,\delta)}}\right]\right) \\
  &\leq & \exp\left(-\frac{1}{2\sigma_T^2(\alpha)}\left[(\mu_1 - \mu_2)-\sqrt{{2\sigma_T^2(\alpha)\beta(T,\delta)}}\right]^2\right).
\end{eqnarray*}
The last inequality follows from Chernoff bound and holds for $T$ such that
$(\mu_1 - \mu_2)>\sqrt{2\sigma^2_T(\alpha)\beta(T,\delta)}.$ Now, for $\gamma
\in (0,1)$ we introduce
\begin{eqnarray*}
  T^*_{\gamma} & := & \inf\left\{t_0 \in \N : \forall t \geq t_0, (\mu_1-\mu_2) - \sqrt{2\sigma^2_t(\alpha)\beta(t,\delta)} > \gamma (\mu_1 - \mu_2)\right\}. 
\end{eqnarray*}
This quantity is well defined as $ \sigma^2_t(\alpha)\beta(t,\delta)$ goes to
zero when $t$ goes to infinity. Then,
\begin{eqnarray*}
  \bE_\nu[\tau]& \leq & T_\gamma^* + \! \sum_{T=T_{\gamma}^*+1}\bP\left(\tau \geq T\right) \\
  &\leq & T_\gamma^* + \! \sum_{T=T_{\gamma}^*+1}\exp\left(-\frac{1}{2\sigma_T^2(\alpha)}\left[(\mu_1 - \mu_2)-\sqrt{{2\sigma_T^2(\alpha)\beta(T,\delta)}}\right]^2\right) \\
  & \leq & T_\gamma^* + \! \sum_{T=T_{\gamma}^*+1}^\infty\exp\left(-\frac{1}{2\sigma_T^2(\alpha)}\gamma^2(\mu_1 - \mu_2)^2\right). 
\end{eqnarray*}
For all $t\in\N^*$, it is easy to show that the following upper bound on
$\sigma_t^2(\alpha)$ holds:
\begin{equation}
  \forall t\in \N, \  \sigma_t^2(\alpha) \leq \frac{(\sigma_1 + \sigma_2)^2}{t}\times \frac{t - \frac{\sigma_1}{\sigma_2}}{t-\frac{\sigma_1}{\sigma_2} -1}.
  \label{ineq:BoundSigma}
\end{equation}
Using the bound (\ref{ineq:BoundSigma}), one has
\begin{eqnarray*}
  \bE_\nu[\tau]& \leq & T_\gamma^* + \int_{0}^{\infty} \exp\left(-\frac{t}{2(\sigma_1 + \sigma_2)^2} \frac{t - \frac{\sigma_1}{\sigma_2}-1}{t - \frac{\sigma_1}{\sigma_2}}\gamma^2(\mu_1-\mu_2)^2\right)dt\\
  & \leq &  T_\gamma^* +  \frac{2(\sigma_1+\sigma_2)^2}{\gamma^2(\mu_1-\mu_2)^2}\exp\left(\frac{\gamma^2(\mu_1-\mu_2)^2}{2(\sigma_1+\sigma_2)^2}\right).
\end{eqnarray*}
We now give an upper bound on $T_\gamma^*$. Let $r\in[0,e/2-1]$. There exists
$N_0(r)$ such that for $t\geq N_0(r)$, $\beta(t,\delta)\leq \log
({t^{1+r}}/{\delta})$.  Using also (\ref{ineq:BoundSigma}), one gets
$T_\gamma^*=\max(N_0(t),\tilde{T}_\gamma)$, where
$$\tilde{T}_\gamma = \inf \left\{ t_0 \in \N : \forall t \geq t_0, \frac{(\mu_1 - \mu_2)^2}{2(\sigma_1+\sigma_2)^2}(1-\gamma)^2 t > \frac{t - \frac{\sigma_1}{\sigma_2}-1}{t - \frac{\sigma_1}{\sigma_2}} \log \frac{t^{1+r}}{\delta}\right\}.$$
If $t > (1 + \gamma \frac{\sigma_1}{\sigma_2})/{\gamma}$ one has $(t -
\frac{\sigma_1}{\sigma_2}-1)/(t - \frac{\sigma_1}{\sigma_2}) \leq
{(1-\gamma)^{-1}}$. Thus $\tilde{T}_\gamma =\max((1 + \gamma
\frac{\sigma_1}{\sigma_2})/{\gamma}, T_\gamma')$, with
$$T_\gamma' = \inf \left\{ t_0 \in \N : \forall t\geq t_0, \exp\left(\frac{(\mu_1 - \mu_2)^2}{2(\sigma_1+\sigma_2)^2}(1-\gamma)^3 t\right) \geq \frac{t^{1+r}}{\delta}\right\}.$$ 
The following Lemma, whose proof can be found below, helps us bound this last
quantity.
\begin{lemma}\label{lem:Garivier2} For every $\beta,\eta>0$ and $s\in[1,e/2]$,
  the following implication is true:
$$x_0 = \frac{s}{\beta}\log\left(\frac{e\log\left({1}/{(\beta^s\eta)}\right)}{\beta^s\eta}\right)  \ \ \ \Rightarrow \ \ \ \forall x\geq x_0, \ \ e^{\beta x} \geq \frac{x^s}{\eta}.$$
\end{lemma}
Applying Lemma \ref{lem:Garivier2} with $\eta=\delta$, $s=1+r$ and $\beta
={(1-{\gamma})^3(\mu_1-\mu_2)^2}/{(2(\sigma_1+\sigma_2)^2)}$ leads to
\[T_\gamma' \leq\frac{(1+r)}{(1-\gamma)^3}\times \frac{2(\sigma_1 +
  \sigma_2)^2}{(\mu_1 - \mu_2)^2}\left[\log \frac{1}{\delta} + \log\log
  \frac{1}{\delta} \right] + R(\mu_1,\mu_2,\sigma_1,\sigma_2,\gamma,r),\] with
\[R(\mu_1,\mu_2,\sigma_1,\sigma_2,\gamma,r)=\frac{1+r}{(1-\gamma)^3}\frac{2(\sigma_1+\sigma_2)^2}{(\mu_1-\mu_2)^2}\left[1
  +
  (1+r)\log\left(\frac{2(\sigma_1+\sigma_2)^2}{(1-\gamma)^3(\mu_1-\mu_2)^2}\right)\right].\]
Now for $\epsilon>0$ fixed, choosing $r$ and $\gamma$ small enough leads to
$$\bE_\nu[\tau] \leq (1+\epsilon)\frac{2(\sigma_1 + \sigma_2)^2}{(\mu_1 - \mu_2)^2}\left[\log \frac{1}{\delta} + \log\log\frac{1}{\delta}\right] + \cC(\mu_1,\mu_2,\sigma_1,\sigma_2,\epsilon),$$
where $\cC$ is a constant independent of $\delta$. It can be noted that $\cC(\mu_1,\mu_2,\sigma_1,\sigma_2,\epsilon)$ goes to infinity when $\epsilon$ goes to zero, but for a fixed $\epsilon>0$, 
\[(1+\epsilon)\frac{2(\sigma_1 + \sigma_2)^2}{(\mu_1 - \mu_2)^2}\log\log\frac{1}{\delta} + \cC(\mu_1,\mu_2,\sigma_1,\sigma_2,\epsilon) = \underset{\delta\rightarrow 0}{o_{\epsilon}}\left(\log \frac{1}{\delta}\right),\]
which concludes the proof.

\medskip

\emph{Proof of Lemma \ref{lem:Garivier2}.} Lemma \ref{lem:Garivier2} easily
follows from the fact that for $\eta>0$ and $s\in [1,e/2]$,
\[x_0 = s \log\left(\frac{e\log\left(\frac{1}{\eta}\right)}{\eta}\right) \ \
\Rightarrow \ \ \ \forall x\geq x_0, \ \ e^x \geq \frac{x^s}{\eta}\] Indeed, it
suffices to apply this statement to $x=x\beta$ and $\eta=\eta\beta^s$. The
mapping $x\mapsto e^x - x^s/\eta$ is increasing when $x\geq s$. As $x_0\geq s$,
it suffices to prove that $x_0$ defined above satisfies $e^{x_0}\geq
x_0^s/\eta$.
\begin{eqnarray*}
  \log\left(\frac{x_0^s}{\eta}\right) & = & s\log\left(s\log\left(\frac{e\log\frac{1}{\eta}}{\eta}\right)\right) + \log\frac{1}{\eta} \\ 
  &=&  s\left(\log(s) + \log \left[\log\frac{1}{\eta} + \log\left(e\log\frac{1}{\eta}\right)\right]\right) +  \log\frac{1}{\eta} \\
  &\leq&  s\left(\log(s) + \log \left[2\log\frac{1}{\eta}\right]\right)  +  \log\frac{1}{\eta}
\end{eqnarray*}
where we use that for all $y$, $\log(y) \leq \frac{1}{e}y$. Then, using that $s\geq 1$,
\begin{eqnarray*}
  \log\left(\frac{x_0^s}{\eta}\right) & \leq & s\left(\log(s) + \log(2) + \log\log\frac{1}{\eta} +  \log\frac{1}{\eta}\right). 
\end{eqnarray*}
For $s\leq \frac{e}{2}$, $\log(s)+\log(2) \leq 1$, hence
\begin{eqnarray*}
  \log\left(\frac{x_0^s}{\eta}\right) & \leq & s\left(1+ \log\log\frac{1}{\eta} +  \log\frac{1}{\eta}\right) = s \log\left(\frac{e\log\left(\frac{1}{\eta}\right)}{\eta}\right) = x_0,
\end{eqnarray*} 
which is equivalent to $e^{x_0} \geq \frac{x_0^s}{\eta}$ and concludes the
proof.

\section{A Refined Exploration Rate for $\alpha$-Elimination\label{proof:DevIneq}}

\subsection{Proof of Theorem~\ref{thm:PACFC}}

According to (\ref{PACExplain}), to prove Theorem \ref{thm:PACFC} it is enough to show that for  
\[\beta(t,\delta)=\log(1/\delta) + 3\log\log(1/\delta) + (3/2)\log\log(et),\]
if $S_t=\sum_{s=1}^tX_s$ is a sum of i.i.d $\norm{0}{1}$ random variables, one has 
\begin{equation}\bP(\exists t\in \N^*: S_t>\sqrt{2 t \beta(t,\delta)})\leq \delta.\label{TOPROVE}\end{equation}
Let $z=\log(1/\delta)$. Using Lemma \ref{thm:subgaussian}, one can write, choosing $x=z+3\log z$ and $\beta=3/2$,
\[\bP\Bigl(\exists t \in \N : S_t > \sqrt{2 t
  \beta(t,\delta)}\Bigr)\leq \frac{\sqrt{e}}{8}\zeta\Bigl(\frac{3}{2}-\frac{3}{4(z+3\log
  z)}\Bigr) \frac{(\sqrt{z + 3\log z} +
  \sqrt{8})^{3/2}}{z^3} \delta.\] 
  It can be shown numerically that for $z \geq 2.03$, 
\[\frac{\sqrt{e}}{8}\zeta\Bigl(\frac{3}{2}-\frac{3}{4(z+3\log
  z)}\Bigr) \frac{(\sqrt{z + 3\log z} +
  \sqrt{8})^{3/2}}{z^3} \leq 1.\]
Thus for $\delta\leq \exp(-2.03) \leq 0.1$, inequality~\eqref{TOPROVE} holds. 

\medskip

\subsection{Proof of Lemma \ref{thm:subgaussian}.} 

We start by stating three technical lemmas, whose proofs are partly omitted.

\begin{lemma}\label{lem:sandwich}
  For every $\eta>0$, every positive integer $k$, and every integer $t$ such
  that $(1+\eta)^{k-1} \leq t \leq (1+\eta)^k$,
  \[\sqrt{\frac{(1+\eta)^{k-1/2}}{t}} + \sqrt{\frac{t}{(1+\eta)^{k-1/2}}} \leq
  (1+\eta)^{1/4} + (1+\eta)^{-1/4}\;.\]
\end{lemma}

\begin{lemma}\label{lem:boundeta}
  For every $\eta>0$,
  \[A(\eta) := \frac{4}{\left((1+\eta)^{1/4} + (1+\eta)^{-1/4}\right)^2} \geq
  1-\frac{\eta^2}{16}.\]
\end{lemma}

\begin{lemma}\label{lem:approxslice}
  Let $t$ be such that $(1+\eta)^{k-1} \leq t \leq (1+\eta)^k$.  Then, 
  \[\sigma\sqrt{2z} \geq \frac{A(\eta)z}{\lambda \sqrt{t}} +
  \frac{\lambda\sigma^2\sqrt{t}}{2}, \ \ \ \text{with} \ \ \ \lambda = \sigma^{-1}\sqrt{2zA(\eta)/(1+\eta)^{k-1/2}}.\]
\end{lemma}
\emph{Proof of Lemma~\ref{lem:approxslice}.}
\[
\frac{A(\eta)z}{\lambda\sqrt{t}} + \frac{\lambda\sigma^2\sqrt{t}}{2} =\frac{\sigma\sqrt{2zA(\eta)}}{2}\left(\sqrt{\frac{(1+\eta)^{k-1/2}}{t}} + \sqrt{\frac{t}{(1+\eta)^{k-1/2}}}\right)\\
\leq \sigma\sqrt{2z}
\]
according to Lemma~\ref{lem:sandwich}.  \qed

An important fact is that for every $\lambda\in\R$, because the $X_i$ are
$\sigma$-subgaussian, $W_t = \exp(\lambda S_t-t\frac{\lambda^2\sigma^2}{2}))$
is a super-martingale, and thus, for every positive $u$,
\begin{equation}
  \bP\left(\bigcup_{t \geq 1} \left\{\lambda S_t-t\frac{\lambda^2\sigma^2}{2} > u \right\}\right) \leq \exp(-u). 
  \label{EquMartingale}
\end{equation}

Let $\eta\in(0, e-1]$ to be defined later, and let $T_k = \N\cap
\left[(1+\eta)^{k-1}, (1+\eta)^{k}\right[$.
\begin{align*}
  &\bP\left(\bigcup_{t\geq 1} \left\{\frac{S_t}{\sigma\sqrt{2t}} > \sqrt{x +
        \beta\log\log(et)} \right\}\right)
  \leq \sum_{k=1}^\infty \bP \left(\bigcup_{t\in T_k} \left\{\frac{S_t}{\sigma\sqrt{2t}} > \sqrt{x + \beta\log\log(et)} \right\}\right)\\
  & \hspace{4cm} \leq \sum_{k=1}^\infty \bP \left(\bigcup_{t\in T_k}
    \left\{\frac{S_t}{\sigma\sqrt{2t}} > \sqrt{x +
        \beta\log\left(k\log(1+\eta)\right)}\right\}\right)\;.
\end{align*}

We use that $\eta\leq e-1$ to obtain the last inequality since this condition
implies $$\log(\log(e(1+\eta)^{k-1})\geq\log(k\log(1+\eta)).$$

For $k\geq 1$, let $z_k = x + \beta\log\left(k\log(1+\eta)\right)$
and $\lambda_k = \sigma^{-1}\sqrt{2z_kA(\eta)/(1+\eta)^{k-1/2}}$.

Lemma~\ref{lem:approxslice} shows that for every $t\in T_k$,
\[\left\{\frac{S_t}{\sigma\sqrt{2t}} > \sqrt{z_k} \right\}
\subset \left\{\frac{S_t}{\sqrt{t}} > \frac{A(\eta)z_k}{\lambda_k \sqrt{t}} +
  \frac{\sigma^2\lambda_k\sqrt{t}}{2}\right\}\;.\]

Thus, by inequality (\ref{EquMartingale}),
\begin{align*}
  \bP \left(\bigcup_{t\in T_k} \left\{\frac{S_t}{\sigma\sqrt{2t}} >
      \sqrt{z_k}\right\}\right)
  &\leq \bP \left(\bigcup_{t\in T_k} \left\{\frac{S_t}{\sqrt{t}} > \frac{A(\eta)z_k}{\lambda_k \sqrt{t}} + \frac{\sigma^2\lambda_k\sqrt{t}}{2}\right\}\right)\\
  & = \bP \left(\bigcup_{t\in T_k} \left\{\lambda_k S_t - \frac{\sigma^2\lambda_k^2 t}{2}> A(\eta)z_k\right\}\right)\\
  &\leq \exp\left(-A(\eta) z_k\right) = \frac{\exp(-A(\eta) x)
  }{(k\log(1+\eta))^{\beta A(\eta)}}\;.
\end{align*}

One chooses $\eta^2 = 8/x$ for $x$ such that $x\geq \frac{8}{(e-1)^2}$ (which
ensures $\eta\leq e-1$). Using Lemma~\ref{lem:boundeta}, one obtains that
$\exp(-A(\eta) x) \leq \sqrt{e} \exp(-x)$. Moreover,
\[
\frac{1}{\log(1+\eta)} \leq \frac{1+\eta}{\eta} =
\frac{\sqrt{x}}{2\sqrt{2}}+1\;.
\]
Thus,
\begin{eqnarray*}\bP \left(\bigcup_{t\in T_k} \left\{\frac{S_t}{\sigma\sqrt{2t}} >
    \sqrt{z_k}\right\}\right) &\leq& \frac{\sqrt{e}}{k^{\beta
    A(\eta)}}\left(\frac{\sqrt{x}}{2\sqrt{2}}+1\right)^{\beta A(\eta)} \exp(-x)
\\ &\leq& \frac{\sqrt{e}}{k^{\beta
    A(\eta)}}\left(\frac{\sqrt{x}}{2\sqrt{2}}+1\right)^{\beta} \exp(-x).\end{eqnarray*} 
Hence,
\begin{align*}
  \bP\left(\bigcup_{t\geq 1}\left\{\frac{S_t}{\sigma\sqrt{2t}} > \sqrt{x + \beta\log\log(et)}\right\}\right) &\leq \sqrt{e}\zeta\left(\beta A(\eta)\right)\left(\frac{\sqrt{x}}{2\sqrt{2}}+1\right)^{\beta A(\eta)}  \exp\left(-x\right)\\
  &\leq \sqrt{e}\zeta\left(\beta
    \left(1-\frac{1}{2x}\right)\right)\left(\frac{\sqrt{x}}{2\sqrt{2}}+1\right)^{\beta}
  \exp\left(-x\right),\end{align*} using the lower bound on $A(\eta)$ given
in Lemma \ref{lem:boundeta} and the fact that $A(\eta)$ is upper bounded by 1.

\section{Bernoulli Bandit Models \label{sec:DetailsBernoulli}}

\subsection{Proof of Lemma \ref{lem:OptBernoulliKL}}

Assume that $\mu_1<\mu_2$. Recall the KL-LUCB algorithm of \cite{COLT13}.  For
two-armed bandit models, this algorithm samples the arms uniformly and builds
for both arms a confidence interval based on KL-divergence
$\cI_a(t)=[l_{a,t/2},u_{a,t/2}]$, with
\begin{eqnarray*}
  u_{a,s}&=&\sup \{q > \hat{\mu}_{a,s} : sd(\hat{\mu}_{a,s},q) \leq \tilde{\beta}(s,\delta)\}, \ \ \text{where} \ \ d(x,y)=\K(\cB(x),\cB(y))\\
  l_{a,s}&=&\inf \ \{q < \hat{\mu}_{a,s} : sd(\hat{\mu}_{a,s},q) \leq \tilde{\beta}(s,\delta)\}, 
\end{eqnarray*}
for some exploration rate that we denote by $\tilde{\beta}(t,\delta)$. The
algorithm stops when the confidence intervals are separated; that is either
$l_{1,t/2}>u_{2,t/2}$ or $l_{2,t/2}>u_{1,t/2}$, and recommends the empirical
best arm.  A picture helps to convince oneself that
\begin{equation}(l_{1,s}>u_{2,s}) \ \Leftrightarrow \
  (\hat{\mu}_{1,s}>\hat{\mu}_{2,s})
  \cap(sd_*(\hat{\mu}_{1,s},\hat{\mu}_{2,s})>\beta(s,\delta))\label{LinkKLLUCB}\end{equation}
Additionally, as mentioned before, $I_*(x,y)$ is very close to the quantity
$d_*(x,y)$ and one has more precisely
$I_*(x,y)<d_*(x,y)$. Using all this, we can upper bound the probability of
error of Algorithm~\ref{AlgoBox:SGLRT} in the following way.
\begin{align*}
  & \bP_\nu\left(\exists t \in 2\N^* :  \hat{\mu}_{1,t/2}>\hat{\mu}_{2,t/2} , tI_*(\hat{\mu}_{1,t/2},\hat{\mu}_{2,t/2})>\beta(t,\delta)\right) \\
  &\hspace{2cm} \leq \bP_\nu\left(\exists t \in 2\N^* : \hat{\mu}_{1,t/2}>\hat{\mu}_{2,t/2} , (t/2)d_*(\hat{\mu}_{1,t/2},\hat{\mu}_{2,t/2})>(\beta(t,\delta)/2)\right) \\
  &\hspace{2cm} = \bP_\nu\left(\exists s \in \N^* : \hat{\mu}_{1,s}>\hat{\mu}_{2,s} , sd_*(\hat{\mu}_{1,s},\hat{\mu}_{2,s})>(\beta(2s,\delta)/2)\right) \\
  &\hspace{2cm} = \bP_\nu(\exists s \in \N^* : l_{1,s}>u_{2,s}) \leq \bP_\nu(\exists s \in \N^* : (\mu_1<l_{1,s}) \cup (\mu_2>u_{2,s})) \\
  &\hspace{2cm} \leq 2\sum_{s=1}^{\infty}\exp(-\beta(2s,\delta)/2)
\end{align*}
where the last inequality follows from an union bound and for example Lemma 4
of \cite{COLT13}. Note that the indices $l_{1,s}$ and $u_{2,s}$ involved here
use the exploration rate $\tilde{\beta}(s,\delta)=\beta(2s,\delta)/2$. The
choice $\beta(t,\delta)$ in the statement of the Lemma shows the last series is
upper bounded by $\delta$, which concludes the proofs.

\subsection{An Asymptotic Bound for the Stopping Time}

\begin{lemma}\label{lem:OptBernoulliKL2} Consider a strategy that uses uniform
  sampling and a stopping rule of the form
  \[\tau = \inf\left\{t\in 2\N^*: tf(\hat{\mu}_{1,t/2},\hat{\mu}_{2,t/2}) \geq
    \log\left(\frac{g(t)}{\delta}\right)\right\}\] where $f$ is a continuous
  function such that $f(\mu_1,\mu_2)\neq 0$ and $g(t)=o(t^r)$ for all
  $r>0$. Then for all $\epsilon>0$,
  \[\bP_\nu\left(\limsup_{\delta\rightarrow 0} \frac{\tau}{\log(1/\delta)} \leq
    \frac{1+\epsilon}{f(\mu_1,\mu_2)}\right)=1.\]
\end{lemma}

\emph{Proof.} We fix $\epsilon>0$ and introduce
\[\sigma = \max \left\{ t\in 2\N^* : f(\hat{\mu}_{1,t/2},\hat{\mu}_{2,t/2})
  \leq \frac{f(\mu_1,\mu_2)}{1+\epsilon/2}\right\}.\] By the law of large
numbers, $\bP(\sigma <+\infty)=1$. Hence, $\lim_{n\rightarrow \infty}\bP(\sigma
\leq n)=1$ and for every $\alpha\in (0,1)$ there exists
$N(\epsilon,\alpha,\mu_1,\mu_2)$ such that $\bP(\sigma \leq
N(\epsilon,\alpha,\mu_1,\mu_2) ) \geq 1 - \alpha$. Therefore, introducing the
event
\begin{eqnarray*}
  E_\alpha & = & \left(\forall t \geq N(\epsilon,\alpha,\mu_1,\mu_2), 
    f(\hat{\mu}_{1,t/2},\hat{\mu}_{2,t/2}) > \frac{f(\mu_1,\mu_2)}{1+\epsilon/2}\right), \ \ \text{one has} \ \ \bP(E_\alpha)\geq 1 - \alpha.
\end{eqnarray*}
On the event $E_\alpha$,
\begin{eqnarray*}
  \tau & \leq & \max\left(N(\epsilon,\alpha,\mu_1,\mu_2) ; \inf\left\{t\in\N : t \frac{f(\mu_1,\mu_2)}{1+\epsilon/2} 
      \geq \log\left(\frac{g(t)}{\delta}\right)\right\}\right) \\
  \tau &  \leq & N(\epsilon,\alpha,\mu_1,\mu_2) + \inf\left\{t\in\N : t \frac{f(\mu_1,\mu_2)}{1+\epsilon/2} 
    \geq \log\left(\frac{g(t)}{\delta}\right)\right\}
\end{eqnarray*}
We can use Lemma \ref{lem:Garivier2} to bound the right term in the right hand
side, which shows that there exists a constant $C(\epsilon,\mu_1,\mu_2)$
independent of $\delta$ such that
\[\tau \leq N(\epsilon,\alpha,\mu_1,\mu_2) +
\frac{1+\epsilon}{f(\mu_1,\mu_2)}\left[\log\frac{1}{\delta} + \log\log
  \frac{1}{\delta}\right] + C(\epsilon,\mu_1,\mu_2) \] Thus we proved that for
all $\alpha > 0$,
\[\bP\left(\limsup_{\delta \rightarrow 0} \frac{\tau}{\log(1/\delta)} \leq
  \frac{1+\epsilon}{f(\mu_1,\mu_2)}\right)\geq 1 - \alpha.\] This concludes the
proof.

\section{Upper and Lower Bounds in the Fixed-Budget Setting}

\subsection{Proof of Theorem \ref{thm:2armsFB}\label{proof:2armsFB}}

Without loss of generality, assume that the bandit model $\nu=(\nu_1,\nu_2)$ is
such that $a^*=1$. Consider any alternative bandit model $\nu'=(\nu_1',\nu_2')$
in which $a^*=2$. Let $\cA$ be a consistent algorithm such that $\tau = t$ and
consider the event $A=(\hat{\cS}_1=1)$. Clearly $A\in\cF_t=\cF_\tau.$

Lemma \ref{lem:Cornerstone} applied to the stopping time $\sigma=t$ a.s. and
the event $A$ gives
\[ \bE_{\nu'}[N_1(t)] \K(\nu'_{1},\nu_{1}) + \bE_{\nu'}[N_2(t)]
\K(\nu'_{2},\nu_{2})\geq d(\bP_{\nu'}(A),\bP_{\nu}(A)).\] Note that
$p_t(\nu)=1-\bP_{\nu}(A)$ and $p_t(\nu')= \bP_{\nu'}(A)$.  As algorithm $\cA$
is correct on both $\nu$ and $\nu'$, for every $\epsilon>0$ there exists
$t_0(\epsilon)$ such that for all $t\geq t_0(\epsilon)$, $\bP_{\nu'}(A) \leq
\epsilon \leq \bP_{\nu}(A)$.  For $t\geq t_0(\epsilon)$,
\[
\bE_{\nu'}[N_1(t)] \K(\nu'_{1},\nu_{1}) + \bE_{\nu'}[N_2(t)]
\K(\nu'_{2},\nu_{2})\geq d(\epsilon,1-p_t(\nu)) \geq (1-\epsilon)\log \frac{1 -
  \epsilon}{p_t(\nu)} + \epsilon \log {\epsilon}.
\]
Taking the limsup and letting $\epsilon$ go to zero, one can show that
\[
\limsup_{t\rightarrow\infty}-\frac{1}{t} \log p_t(\nu) \leq
\limsup_{t\rightarrow\infty}\sum_{a=1}^2\frac{\bE_{\nu'}[N_a(t)]}{t}
\K(\nu'_{a},\nu_{a}) \leq \max_{a=1,2}\K(\nu_{a}',\nu_{a}).
\]
Optimizing over the possible model $\nu'$ satisfying $\mu_1'<\mu_2'$ to make
the right hand side of the inequality as small as possible gives the result.

For algorithms using uniform sampling, $\limsup -\frac{1}{t}\log p_t(\nu)$ is
upper bounded by the quantity $(\K(\nu'_{1},\nu_{1}) +\K(\nu'_{2},\nu_{2}))/2$, which yields
the second statement of the Theorem.

\subsection{An Optimal Static Strategy for Exponential
  Families \label{proof:ConcExp}}

Bounding the probability of error of a static strategy using $n_1$ samples from
arm 1 and $n_2$ samples from arm 2 relies on the following lemma.

\begin{lemma} \label{lem:ConcExp} Let $(X_{1,t})_{t\in\N}$ and
  $(X_{2,t})_{t\in\N}$ be two independent i.i.d sequences, such that $X_{1,1}
  \sim \nu_{\theta_1}$ and $X_{2,1}\sim \nu_{\theta_2}$ belong to an
  exponential family. Assume that $\mu(\theta_1) > \mu(\theta_2)$. Then
  \begin{equation}\bP\left(\frac{1}{n_1}\sum_{t=1}^{n_1}X_{1,t}<
      \frac{1}{n_2}\sum_{t=1}^{n_2}X_{2,t}\right) \leq \exp(-(n_1 + n_2)
    g_\alpha(\theta_1,\theta_2)),\label{MainIneq}\end{equation}
  where $\alpha = \frac{n_1}{n_1 + n_2}$ and $g_{\alpha}(\theta_1,\theta_2)  := \alpha \Kb(\alpha\theta_1 + (1-\alpha)\theta_2,\theta_1) + (1-\alpha)\Kb(\alpha \theta_1 + (1-\alpha)\theta_2 , \theta_2).$
\end{lemma}
The function $\alpha \mapsto g_\alpha(\theta_1,\theta_2)$, can be maximized
analytically, and the value $\alpha^*$ that realizes the maximum is given by
\begin{eqnarray*}
  \Kb(\alpha^*\theta_1 + (1-\alpha^*)\theta_2,\theta_1) & = & \Kb(\alpha^*\theta_1 + (1-\alpha^*)\theta_2,\theta_2) \\
  \alpha^* \theta_1 + (1- \alpha^*)\theta_1 & = & \theta^* \\
  \alpha^* &=&\frac{\theta^* - \theta_2}{\theta_1 - \theta_2}
\end{eqnarray*}
where $\theta^*$ is defined by
$\Kb(\theta^*,\theta_1)=\Kb(\theta^*,\theta_2)=\Kb^*(\theta_1,\theta_2)$.  More
interestingly, the associated rate is such that
\[
g_{\alpha^*}(\theta_1,\theta_2)=\alpha^*\Kb(\theta^*,\theta_1) +
(1-\alpha^*)\Kb(\theta^*,\theta_2) =\Kb^*(\theta_1,\theta_2),
\]
which leads to Theorem \ref{prop:Bernoulli1}.

\begin{remark}\label{rem:Match} When $\mu_1>\mu_2$, applying Lemma
  \ref{lem:ConcExp} with $n_1=n_2=t/2$ yields
  \[\bP\left(\hat{\mu}_{1,t/2}<\mu_{2,t/2}\right) \leq
  \exp\left(-\frac{\Kb\left(\theta_1,\frac{\theta_1 +\theta_2}{2}\right)
      +\Kb\left(\theta_2,\frac{\theta_1 +\theta_2}{2}\right)}{2}\,t\right) =
  \exp\big(-I_*(\nu)t\big),\] which shows that the strategy using uniform sampling and recommending the empirical best arm matches the
  lower bound (\ref{BoundUnifFB}) in Theorem \ref{thm:2armsFB}.
\end{remark}

\emph{Proof of Lemma \ref{lem:ConcExp}.} The i.i.d. sequences $(X_{1,t})_{t\in\N}$ and $(X_{2,t})_{t\in\N}$ have 
respective densities $f_{\theta_1}$ and $f_{\theta_2}$ where
$f_\theta(x)=\exp(\theta x - b(\theta))$ and $\mu(\theta_1)=\mu_1,
\mu(\theta_2)=\mu_2$. $\alpha$ is such that $n_1=\alpha n $ and
$n_2=(1-\alpha)n$. One can write
\[
\bP\left(\frac{1}{n_1}\sum_{t=1}^{n_1}X_{1,t}-
  \frac{1}{n_2}\sum_{t=1}^{n_2}X_{2,t}< 0\right) = \bP\left(\alpha
  \sum_{t=1}^{n_2}X_{2,t} - (1-\alpha)\sum_{t=1}^{n_1}X_{1,t} \geq 0 \right).
\]
For every $\lambda > 0$, multiplying by $\lambda$, taking the exponential of
the two sides and using Markov's inequality (this technique is often referred
to as Chernoff's method), one gets
\begin{multline*}
  \bP\left(\frac{1}{n_1}\sum_{t=1}^{n_1}X_{1,t}-
    \frac{1}{n_2}\sum_{t=1}^{n_2}X_{2,t}< 0\right) \leq
  \left(\bE_\nu[e^{\lambda \alpha X_{2,1}}]\right)^{(1-\alpha) n}
  \left(\bE_\nu[e^{\lambda (1-\alpha )X_{1,1}}]\right)^{\alpha n} \\
  = \exp\biggl(n \underbrace{\left[(1-\alpha) \phi_{X_{2,1}}(\lambda \alpha) +
      \alpha
      \phi_{X_{1,1}}(-(1-\alpha)\lambda)\right]}_{G_\alpha(\lambda)}\biggr)
\end{multline*}
with $\phi_X(\lambda)=\log \bE_\nu[e^{\lambda X}]$ for any random variable
$X$. If $X \sim f_\theta$ a direct computation gives $\phi_X(\lambda)=b(\lambda
+ \theta) - b(\theta)$. Therefore the function $G_\alpha(\lambda)$ introduced
above rewrites
\[G_\alpha(\lambda) = (1-\alpha)(b(\lambda \alpha + \theta_2) - b(\theta_2)) +
\alpha(b(\theta_1 - (1-\alpha)\lambda) - b(\theta_1)).\] Using that
$b'(x)=\mu(x)$, we can compute the derivative of $G$ and see that this function
as a unique minimum in $\lambda^*$ given by
\[\mu(\theta_1-(1-\alpha)\lambda^*) = \mu(\theta_2 + \alpha\lambda^*) \ \
\Leftrightarrow \ \ \theta_1-(1-\alpha)\lambda^* = \theta_2 + \alpha\lambda^* \
\ \Leftrightarrow \ \ \lambda^* = \theta_1-\theta_2,
\]
using that $\theta \mapsto \mu(\theta)$ is one-to-one. One can also show that
\begin{eqnarray*}
  G(\lambda^*)& = &(1-\alpha)[ b(\alpha \theta_1 +(1-\alpha)\theta_2) - b(\theta_2)] + \alpha[b(\alpha \theta_1 + (1-\alpha)\theta_2) - b(\theta_1)]. 
\end{eqnarray*}
Using the expression of the KL-divergence between $\nu_{\theta_1}$ and
$\nu_{\theta_2}$ as a function of the natural parameters:
$\Kb(\theta_1,\theta_2)= \mu(\theta_1)(\theta_1 - \theta_2) - b(\theta_1) +
b(\theta_2)$, one can also show that
\begin{align*}
  & \alpha \Kb(\alpha \theta_1 + (1-\alpha)\theta_2 , \theta_1) \\
  & \hspace{2cm} =  -\alpha(1-\alpha)\mu(\alpha\theta_1 + (1-\alpha)\theta_2)(\theta_1 - \theta_2) + \alpha[-b(\alpha \theta_1 + (1-\alpha)\theta_2) + b(\theta_1)] \\
  & (1-\alpha) \Kb(\alpha \theta_1 + (1-\alpha)\theta_2 , \theta_2) \\
  & \hspace{2cm}= \alpha(1-\alpha)\mu(\alpha\theta_1 +
  (1-\alpha)\theta_2)(\theta_1 - \theta_2) + (1-\alpha)[-b(\alpha \theta_1 +
  (1-\alpha)\theta_2) + b(\theta_2)]
\end{align*}
Summing these two equalities leads to
$$G(\lambda^*) = -\left[\alpha \Kb(\alpha \theta_1 + (1-\alpha)\theta_2 , \theta_1)  +  (1-\alpha) \Kb(\alpha \theta_1 + (1-\alpha)\theta_2 , \theta_2)\right] = - g_\alpha(\theta_1,\theta_2). $$
Hence the inequality $\bP\left(\frac{1}{n_1}\sum_{t=1}^{n_1}X_{1,t}<
  \frac{1}{n_2}\sum_{t=1}^{n_2}X_{2,t}\right)\leq \exp(nG(\lambda^*))$
concludes the proof.

\subsection{Proof of Theorem~\ref{thm:FBGene2}\label{proof:FBGene2}}

First, with $\Delta_a$ as defined in the introduction, there exists one arm
$a\in\{1,\dots,K\}$ such that $\bE_\nu[N_a(t)]\leq {2\sigma^2
  t}/{(H(\nu)\Delta_a^2)}.$ Otherwise, a contradiction is easily obtained.

\medskip
  
\emph{Case 1} If $a\in \{1,\dots m\}$ there exists $b\in \{m+1,\dots K\}$
such that $\bE_\nu[N_b(t)]\leq \frac{2\sigma^2t}{H^-(\nu)\Delta_b^2}$.

\smallskip

\emph{Case 2} If $a\in \{m+1,\dots K\}$ there exists $b\in \{1,\dots, m\}$
such that $\bE_\nu[N_b(t)]\leq \frac{2\sigma^2t}{H^+(\nu)\Delta_b^2}$.

These two cases are very similar, and the idea is to propose an easier
alternative model in which we change only arm $a$ and $b$, the arms that are
less drawn among the set of good and the set of bad arms. Assume that we are in
Case 1.  We introduce $\nu^{[a,b]}$ a Gaussian bandit model such that:
\begin{equation*}
  \left\{
    \begin{array}{ccl}
      \mu_k'&=&\mu_k \ \text{for all} \ k\notin\{a,b\} \\
      \mu_a'&=&\mu_a - 2\Delta_b \\
      \mu'_b&=&\mu_b + 2\Delta_a
    \end{array}
  \right.
\end{equation*}
In $\nu^{[a,b]}$ good arm $a$ becomes a bad arm and bad arm $b$ becomes a good
arm. One can easily check (or convince oneself with Figure
\ref{fig:Illustration}) that $H(\nu^{[a,b]})\leq H(\nu)$ and as already
explained,  $\nu$ and $\nu^{[a,b]}$ do not share their optimal arms.  Thus Lemma \ref{lem:CornerVariant} yields
\begin{eqnarray*}
  \max\left(p_t(\nu),p_t(\nu^{[a,b]})\right) &\geq & \frac{1}{4}\exp\left(-\left[\bE_\nu[N_a(t)]\K(\nu_{a},\nu'_{a}))+
      \bE_\nu[N_b(t)]\K(\nu_{b},\nu'_{b})\right]\right) \\
  &=&  \frac{1}{4}\exp\left(-\left[\bE_\nu[N_a(t)]\frac{(2\Delta_a)^2}{2\sigma^2} +\bE_\nu[N_b(t)]\frac{(2\Delta_b)^2}{2\sigma^2}\right] \right), \end{eqnarray*}
thus  
\begin{eqnarray*}
   \max\left(p_t(\nu),p_t(\nu^{[a,b]})\right)  & \geq & \frac{1}{4}\exp\left(-\left[\frac{2\sigma^2t}{H\Delta_a^2}\frac{4\Delta_a^2}{2\sigma^2} + \frac{2\sigma^2t}{H^-\Delta_b^2}\frac{4\Delta_b^2}{2\sigma^2}\right]\right) \\
  & = & \frac{1}{4}\exp\left(-\frac{4t}{\tilde{H}}\right) \ \ \text{with} \ \ \tilde{H}=\frac{H H^-}{H + H^-}.
\end{eqnarray*}

\bibliography{biblioTech}

\end{document}